\documentclass[Afour,sageh,times]{sagej}

\setcitestyle{numbers}

\usepackage{enumitem}
\usepackage{microtype}

\usepackage{moreverb,url}
\usepackage[colorlinks,bookmarksopen,bookmarksnumbered,citecolor=red,urlcolor=red]{hyperref}

\usepackage[dvipsnames, rgb]{xcolor}
\usepackage{tikz}
\usetikzlibrary{shapes}

\usepackage[acronym]{glossaries}
\usepackage{xspace}

\usepackage{amsmath}

\usepackage[ruled,linesnumbered,vlined]{algorithm2e}

\usepackage{tabularx}
\usepackage{arydshln}
\usepackage{etoolbox}
\usepackage{longtable}
\usepackage{pdflscape}
\usepackage{colortbl}
\usepackage{booktabs}
\usepackage{multirow}

\newcommand\BibTeX{{\rmfamily B\kern-.05em \textsc{i\kern-.025em b}\kern-.08em
T\kern-.1667em\lower.7ex\hbox{E}\kern-.125emX}}

\makeatletter

\renewcommand{\@maketitle}{%
\if@Royal\vspace*{-20pt}\fi
\if@Crown\vspace*{-20pt}\fi
\vspace*{-34pt}%
\null%
\begin{center}
\begin{sf}

\begin{minipage}[t]{\textwidth}
  \vskip 12.5pt%
  {\raggedright\titlesize\textbf{\@title}\par}%
  \vskip 1.5em%
\end{minipage}
\vspace{10pt}
{\par\large
\raggedright\textbf{\@author}\par}

\vskip 20pt%
{\noindent\usebox\absbox\par}
{\vspace{20pt}\noindent\normalsize\@keywords\par}

\end{sf}
\end{center}
\vspace{22pt}
}

\newcommand{\reducedplus}{\mathpalette\reduced@plus\relax}
\newcommand{\reduced@plus}[2]{%
  \sbox6{$\m@th#1+$}%
  \sbox8{\scalebox{0.875}{\copy6}}%
  \dimen@=\dimexpr(\wd6-\wd8)/3\relax
  \raisebox{\dimen@}{\box8}%
}
\newcommand{\boxoperation}[2][\mathbin]{%
  #1{\mathpalette\box@operation{#2}}%
}
\newcommand{\box@operation}[2]{%
  \ooalign{$\m@th#1\boxempty$\cr\hidewidth$\m@th#1#2$\hidewidth\cr}%
}

\newcommand{\etalcite}[1]{et al.~\cite{#1}}
\newcommand{\ie}{i.e.,\xspace}

\newcommand\method{TACO\xspace}

\def\secref#1{Sec.~\ref{#1}}
\def\figref#1{Fig.~\ref{#1}}
\def\tabref#1{Tab.~\ref{#1}}
\renewcommand{\eqref}[1]{Eq.~(\ref{#1})}
\def\algref#1{Alg.~\ref{#1}}

\SetKwIF{Upon}{}{}{upon}{then}{}{}{}

\DeclareMathOperator{\argmin}{\underset{\textbf{x}}{\mathrm{argmin}}}

\tikzset{cross/.style={cross out, draw=black, minimum size=10pt, inner sep=0.5pt, outer sep=0.5pt}, cross/.default={5pt}}

\newcommand{\pinkcircle}{\raisebox{0pt}{\tikz{\node[draw,scale=0.6,circle,fill=pink!90](){};}}}
\newcommand{\greentriangle}{\raisebox{0pt}{\tikz{\node[draw,scale=0.4,regular polygon, regular polygon sides=3,fill=OliveGreen!,rotate=0](){};}}}

\newcommand{\browntriangle}{\raisebox{0pt}{\tikz{\node[draw,scale=0.4,regular polygon, regular polygon sides=3,fill=RawSienna!70!,rotate=180](){};}}}

\definecolor{mplSkyBlue}{RGB}{135,206,235}
\newcommand{\lightbluesquare}{\raisebox{0pt}{\tikz{\node[draw,scale=0.5,regular polygon, regular polygon sides=4,fill=mplSkyBlue](){};}}}

\newcommand{\goldendiamond}{\raisebox{0pt}{\tikz{\node[draw,scale=0.5,diamond,fill=YellowOrange!70!](){};}}}

\newcommand{\bluehexa}{\raisebox{0pt}{\tikz{\node[draw,scale=0.6,regular polygon, regular polygon sides=6,fill=black!20!blue](){};}}}
\newcommand{\redx}{\raisebox{-2pt}{\tikz{\node[draw,scale=0.5,cross,rotate=0,draw=red, line width=1mm=5mm](){};}}}

\newcommand{\purplecross}{\raisebox{-2pt}{\tikz{\node[draw,scale=0.5,cross,rotate=45,draw=blue!60!orange, line width=1mm=5mm](){};}}}

\makeatother

\begin{document}

\runninghead{Olivastri et al.}

\title{\method: A Test and Check Framework for Robust Pose Graph Optimization}

\author{Emilio Olivastri\affilnum{1}, Alberto Pretto\affilnum{2}, and Tobias Fischer\affilnum{1}}

\affiliation{\affilnum{1}Queensland University of Technology, AU\\
\affilnum{2}University of Padova, IT}

\corrauth{Emilio Olivastri, Queensland University of Technology,
QUT Centre for Robotics,
Brisbane, Queensland,
4000, AU.}

\email{e.olivastri@qut.edu.au}

\newacronym{ATE}{ATE}{Absolute Trajectory Error}
\newacronym{MAP}{MAP}{Maximum a Posteriori}
\newacronym{NLLS}{NLLS}{Non-Linear Least Squares}
\newacronym{PGO}{PGO}{Pose Graph Optimization}
\newacronym{RPGO}{RPGO}{Robust Pose Graph Optimization}
\newacronym{SE}{SE}{Special Euclidean Group}
\newacronym{SLAM}{SLAM}{Simultaneous Localization and Mapping}
\newacronym{TACO}{TACO}{Test And Check Optimization}
\newacronym{IPC}{IPC}{Incremental Probabilistic Consensus}
\newacronym{SOS}{SOS}{Switchable Outlier Sanitization}

\begin{abstract}
\acrlong{PGO} (\acrshort{PGO}) is one of the most widely adopted approaches for solving Simultaneous Localization and Mapping (\acrshort{SLAM}) problems. However, \acrshort{PGO} approaches are particularly sensitive to outliers, which can substantially degrade the quality of the estimated trajectories. These outliers arise from incorrect place recognition associations caused by perceptual aliasing in the environment. In this paper, we present \acrshort{TACO} (short for \acrlong{TACO}), a robust optimization framework designed to filter out outliers from \acrshort{PGO} systems. Rather than explicitly modeling measurements as inliers or outliers, \acrshort{TACO} finds an approximation to the maximally consistent set of measurements incrementally through two complementary components: (i) The \textit{test} component, namely the \acrlong{IPC} (\acrshort{IPC}) algorithm, evaluates the consistency of each incoming loop closure online. (ii) The \textit{check} component dubbed \acrlong{SOS} leverages the existing Switchable Constraints to periodically sanitize any inconsistent measurements from the consistent set that \acrshort{IPC} may have mistakenly included. We evaluate \acrshort{TACO} on 2D SLAM and 3D Visual SLAM datasets against several state-of-the-art methods. The results show robustness comparable to state-of-the-art offline methods while preserving the computational efficiency required for online deployment, achieving a success rate above $90\%$ in 2D and $83\%$ in 3D across outlier rates up to 50\%, with mean convergence times of approximately 45\,ms and 100\,ms, respectively. We release an open-source implementation of our method with this paper.
\end{abstract}


\maketitle

\section{Introduction}
\begin{figure*}[t!]
    \centering
    \includegraphics[width=1.0\textwidth]{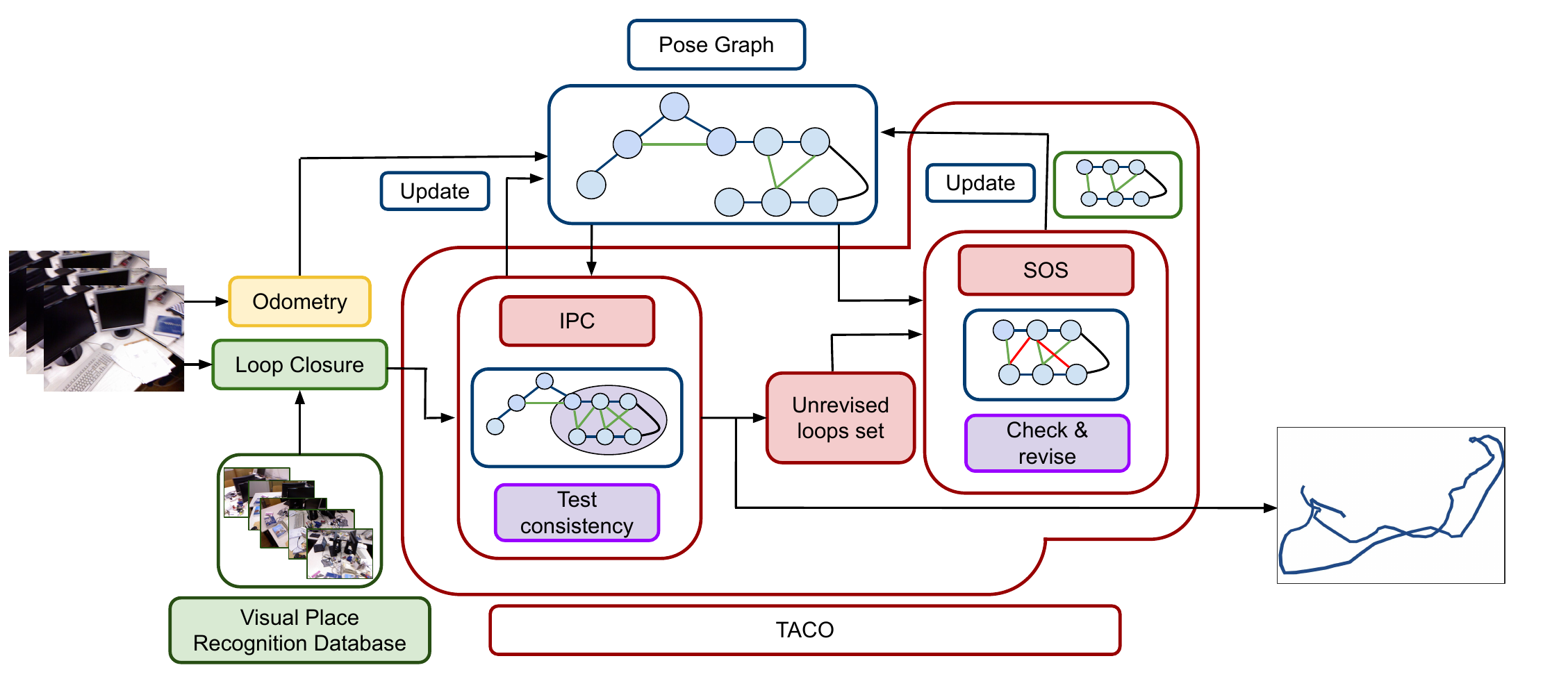}
    \caption{Overview of the \acrshort{TACO} framework used inside a Visual \acrshort{SLAM} system. The front-end generates both odometry and loop closures constraints from a sequence of images. \acrshort{IPC} tests the consistency of candidate loop closures against their best loop matches. The output of \acrshort{IPC} provides an online estimate of the trajectory (bottom right in blue). Each candidate loop closure, classified as an inlier, is then added to a set called \emph{unrevised set}. Once such a set reaches size $M$, \acrshort{SOS} will double-check the loop closures and update the original pose graph, along with the estimates of the related variables.}
    \label{fig:teaserTACO}
\end{figure*}

A widely used approach for solving the \acrlong{MAP} (\acrshort{MAP}) problem in \acrlong{SLAM} (\acrshort{SLAM}) is \emph{\acrlong{PGO}}~\cite{Cadena16tro-SLAMfuture, 5681215}, where the problem is formulated as a graph whose nodes represent robot poses and whose edges encode spatial constraints derived from front-end measurements.

Within a \acrshort{SLAM} pipeline, the \emph{front-end} is responsible for constructing these constraints, including relative motion estimates between consecutive sensor observations (referred to as \emph{odometry})~\cite{6096039, lee2024lidar}, and the detection of previously visited locations~\cite{10261441, luo20243d}, commonly referred to as \emph{place recognition} or \emph{loop closure detection}. The \acrshort{PGO} back-end then estimates the robot trajectory using \acrlong{NLLS} (\acrshort{NLLS}) techniques.

Despite their widespread use~\cite{grisetti2011g2o, gtsam, Agarwal_Ceres_Solver_2022}, \acrshort{NLLS} methods are inherently sensitive to initialization and to the presence of outliers due to the non-convexity of the underlying problem~\cite{burke1995gauss, more2006levenberg, nocedal1999numerical}. Our work addresses the challenge of handling outliers within the \acrshort{NLLS} optimization process, where even a single outlier can invalidate the estimated trajectory. In this context, an \emph{outlier} refers to a spurious loop closure that incorrectly associates distinct places, whereas an \emph{inlier} corresponds to a correct association between observations of the same location.

Most \acrshort{SLAM} systems~\cite{rosen2019se, engelLSD, 9440682, 8593376, guadagnino2025kiss} adopt conservative thresholds in the loop closure detector to minimize the likelihood of introducing outliers. However, in complex environments, perceptual aliasing makes it unrealistic to expect the front-end to avoid such errors when relying on isolated observations. Consequently, it should be the back-end that takes charge of proactively dealing with erroneous measurements, thanks to its ability to access the complete measurement history.

Existing approaches extend classical \acrshort{NLLS} solvers by incorporating robust cost functions~\cite{barron2019general, agarwal2013robust}, performing constraint consistency checks~\cite{matas2004randomized, mangelson2018pairwise, lusk2021clipper, neira2001data}, or adopting graduated strategies that progressively reintroduce non-convexity~\cite{yang2020graduated, mcgann2023robust}. However, these approaches often increase sensitivity to initialization or are primarily suited for offline processing.

To address these limitations, we introduce \acrshort{TACO}, short for \acrlong{TACO}, a framework for \acrlong{RPGO} (\acrshort{RPGO}). \acrshort{TACO} consists of two complementary components: a ``test'' and a ``check'' component (see \figref{fig:teaserTACO} for an overview). The test component, namely the \acrlong{IPC} (\acrshort{IPC}) algorithm, falls under the category of Consistent Set Maximization methods. Its goal is to approximate, in an incremental manner, the combinatorial problem of identifying the maximally consistent set of measurements, therefore enabling online performance.

The check component, namely \acrlong{SOS} (\acrshort{SOS}), leverages the Switchable Constraints (SC) framework~\cite{sunderhauf2012switchable} to periodically filter out, or ``\emph{sanitize}", outliers from the consistent set of measurements that were mistakenly included by \acrshort{IPC}. Using SC strictly for consistency evaluation rather than full inlier-outlier classification allows effective sanitization with limited temporal overhead.

We also highlight the limitations of standard \acrshort{SLAM} evaluation metrics for assessing robustness in the presence of outliers, and propose metrics specifically designed to evaluate the performance of \acrshort{RPGO}. Experimental results on both 2D and 3D Visual \acrshort{SLAM} benchmarks show that our proposed \acrshort{TACO} attains robustness comparable to state-of-the-art offline methods, while preserving the fast responsiveness required for an online \acrshort{SLAM} system. This reduces reliance on outdated estimates during integration of new data without compromising robustness, which is essential for navigating in challenging environments.

The primary contributions of this work can be summarized as follows:
\begin{enumerate}[noitemsep,topsep=0pt]
    \item The \acrlong{TACO} Framework: A two-stage framework that combines incremental consensus (\acrshort{IPC}) with a dedicated error recovery module (\acrshort{SOS}), effectively combining the advantages of online and offline approaches;
    \item A discussion of the assumptions and heuristics from the robust optimization literature that guide the design of the framework;
    \item An analysis highlighting the limitations of traditional SLAM evaluation metrics for assessing robustness, together with the identification of more suitable metrics for evaluating \acrshort{RPGO} performance.
\end{enumerate}    
An open-source implementation of \acrshort{TACO} is released with this work\footnote{\url{https://github.com/EmilioOlivastri/TACO}}.

The remainder of this paper is organized as follows: \secref{sec:rel_works} reviews related work on robust optimization and outlier rejection. \secref{sec:preliminaries} provides the necessary background and foundational concepts for the paper. In \secref{sec:methodology}, we introduce the \acrshort{TACO} framework and its components, \acrshort{IPC} and \acrshort{SOS}. A discussion about suitable metrics for \acrlong{RPGO} (\acrshort{RPGO}) is reported in \secref{sec:metrics}, while \secref{sec:experiments} presents a comprehensive evaluation of the proposed method, comparing it against state-of-the-art solutions on datasets derived from 2D and 3D Visual SLAM real-world data, specifically adapted for RPGO problems. Finally, \secref{sec:conclusions} summarizes our findings and draws conclusions.

This paper extends our previous conference version presented at ICRA~\cite{10611214}, which introduced the \acrshort{IPC} algorithm as a standalone framework. In this work, we extend that formulation by introducing the \acrshort{SOS} component, which addresses a key limitation of \acrshort{IPC} by actively revising accepted loop closures and recovering from degenerate estimates. Furthermore, we propose a set of evaluation metrics for assessing robust pose graph optimization frameworks effectively. Finally, we expand the experimental evaluation from 2D datasets to include 3D Visual SLAM benchmarks, and provide ablation studies analyzing the effect of key parameters on the behavior and performance of the proposed framework.
\section{Related Work}
\label{sec:rel_works}
Interest in robust pose graph optimization has increased in recent years~\cite{carlone2025slam}, as optimization is a delicate process, and even a few outliers can substantially impact the final solution. To mitigate this, various techniques for robust pose graph optimization have been proposed throughout the literature.

\subsection{M-Estimators}
One of the earliest approaches to mitigate the effects of outliers in the optimization process is the use of M-estimators~\cite{black1996unification}. These robust cost functions replace the standard quadratic error to prevent the optimizer from minimizing errors associated with outliers. M-estimators, such as Huber, Cauchy, Geman-McClure, and Truncated-Least-Squares, reduce or nullify the influence of errors exceeding a maximum acceptable threshold. The region within which errors are considered is referred to as the ``trust region". M-estimators inject additional non-linearity into the problem and are sensitive to the threshold defining the trust region.

Selecting an appropriate threshold and M-estimator is problem-specific and often non-trivial. 
Barron~\etalcite{barron2019general} introduce a new general M-estimator called ADAPT. The ADAPT robust cost function is capable of representing the various robust cost functions, selecting the preferred one through a control parameter. This approach eliminates the need to manually choose the most suitable cost function, as the optimizer treats the control parameter as a latent variable and automatically determines the appropriate level of robustness.
In contrast, \method avoids introducing additional non-linearities to reduce sensitivity to problem initialization, which M-estimator-based approaches typically suffer from.

\subsection{Graduated Non-Convexity}
Graduated Non-Convexity (GNC)~\cite{yang2020graduated} is an iterative framework that solves a non-convex problem by first addressing its more convex approximations. Initially, a convex approximation of the original problem is solved, and with each iteration, non-convexity is gradually introduced until the original problem is fully restored. The core idea behind GNC is that the solution to a simpler version of the problem serves as an effective starting point for solving the more complex problem. GNC filters out outliers more effectively than the direct application of an M-estimator. This arises from GNC's ability to better manage the non-linearity introduced by robust cost functions, albeit at the expense of increased convergence time. 

Ramezani~\etalcite{ramezani2022aeros} aim to address both the non-linearity introduced by robust cost functions and the task of manually selecting a cost function with AEROS. AEROS combines the GNC paradigm with the general ADAPT loss~\cite{barron2019general}. 
McGann~\etalcite{mcgann2023robust} propose a version of GNC suited for incremental solvers such as iSAM2~\cite{kaess2012isam2}. The use of the Scale-Invariant Graduated (SIG) Kernel enables a dramatic reduction in the number of iteration steps required to achieve convergence. However, Graduated Non-Convexity-based approaches remain more suitable for offline processing, whereas \method is designed for online deployment.

\subsection{Augmented Pose Graph Optimization}
A widely adopted alternative for solving robust pose graph optimization involves defining an augmented problem that integrates outlier removal as part of the cost function to be minimized. In~\cite{sunderhauf2012switchable}, ``switchable" variables are introduced to the classic PGO formulation. These variables are used to determine whether a measurement is an inlier or an outlier, automatically removing outliers from the optimization process. Agarwal~\etalcite{agarwal2013robust} present an equivalent formulation that eliminates the need for these additional variables, thereby reducing convergence time. Instead of using variables to model the presence of outliers, Olson~\etalcite{olson2013inference} employ a multi-modal Gaussian distribution. The optimizer selects the most promising Gaussian at each iteration by using the max-mixture distribution as the underlying model. In \method, the use of switchable variables is restricted to control the increase in problem complexity, enabling a real-time revision step.

\subsection{Maximal Consistent Set of Measurements}
\label{subsec:maximalset}
As noted in~\cite{carlone2014selecting}, the inlier or outlier status of measurements is not always directly observable. Consequently, joint consistency among measurements has emerged as a key criterion for robust PGO. Instead of explicitly distinguishing inliers from outliers, methods in this category prioritize identifying the largest subset of mutually consistent measurements, leveraging the inherent consistency between inliers. However, determining the maximally consistent set of measurements poses a problem of combinatorial complexity.

This has pushed the research community to develop effective heuristics and approximations for this problem. Neira~\etalcite{neira2001data} propose an exact method that is not exponential in most cases, but it is still not viable for real-time or online applications. The proposed Joint Compatibility Branch and Bound (JCBB) method leverages the Branch and Bound technique in conjunction with the joint compatibility test between measurements.
Mangelson~\etalcite{mangelson2018pairwise} propose the Pairwise Consistency Maximization (PCM) method. PCM approximates the joint consistency check using pairwise consistency and employs heuristic Max-Clique algorithms to compute the largest set of pairwise consistent measurements.

ROBIN~\cite{shi2021robin} introduces and adopts a new \emph{invariance} criterion to effectively build the compatibility graph. An invariant assesses the consistency between measurements independently of the problem's state variable estimates. Defining an effective invariant requires exploiting the inherent priors and properties of the specific problem at hand. Lusk~\etalcite{lusk2021clipper} address different registration problems and define specific geometric-based invariants for each. Their proposed method uses the projection gradient ascent with a back-tracking line search to maintain low execution time while still being able to compute the largest set of consistent measurements.

A more heuristic approach is proposed in~\cite{latif2013robust}, where measurements are temporally grouped into clusters, and these clusters are used for testing both intra- and inter-cluster consistency.  The largest set of consistent measurements is approximated to all measurements that pass both tests. Wu~\etalcite{wu2020cluster} build on the previous approach by introducing an additional term designed to mitigate inconsistencies caused by noisy odometry measurements. However, it comes at the cost of increased execution time. In contrast to these approaches, which operate on the full pose graph and repeatedly reprocess all measurements at each iteration, making them inherently offline, \method limits optimization to smaller subgraphs identified by the most recent loop closure to enable efficient online operation.
\section{Preliminaries}
\label{sec:preliminaries}
Throughout this work, lowercase letters $x$ denote scalar quantities, while bold lowercase letters $\mathbf{x}$ denote vectors. Bold uppercase letters $\mathbf{X}$ represent matrices, and calligraphic uppercase letters $\mathcal{X}$ are used for sets. Subscripts indicate discrete time indices. Finally, $SE(d)$ denotes the \acrlong{SE} of dimension $d$.

\subsection{System Setup}
\label{sec:system_setup}
We expect the robot to be equipped with (i) an ego-motion estimation front-end~\cite{zhang2014loam, wang2017stereo} that provides, over time, the rigid transformations between consecutive poses $\mathbf{x}_i, \mathbf{x}_{i+1} \in \acrshort{SE}(d)$ traversed by the robot (\ie an \emph{odometry measurement} or \emph{constraint}); (ii) a place recognition module~\cite{6202705, lim2025kiss, schubert2023visual} that can detect medium- and long-term previously seen locations and estimate the transformation $\mathbf{z}_{i,j} \in \acrshort{SE}(d)$ between the related previous pose $\mathbf{x}_i$ and the current one $\mathbf{x}_j$ (\ie a \emph{loop closure measurement} or \emph{constraint}). Here, $d \in \{2, 3\}$ depending on whether the robot operates in two or three dimensions. These poses represent the robot's poses in the map's coordinate system.

\subsection{Pose Graph Optimization Formulation}
The prevailing paradigm for back-end optimization is called \acrlong{PGO} (\acrshort{PGO}), which employs a graphical representation where nodes represent the robot’s poses, and edges capture the spatial relationships between them. This graphical representation provides a compact and expressive way to formulate and solve the underlying optimization problem. Here, we introduce the core concepts of \acrshort{PGO}.

A pose graph $\mathcal{G} = (\mathcal{V},\mathcal{E})$ uses nodes in $\mathcal{V}$ to represent the discretized poses $\mathbf{x}_i$ of the robot, while the edges $(i,j) \in \mathcal{E}$ encode the spatial relation between the nodes $\mathbf{x}_i$ and $\mathbf{x}_j$.
The full state can be represented using the vector $\mathbf{x}= [\mathbf{x}_1, \dots, \mathbf{x}_n]$, where the subscript defines the temporal order and $n$ is the total number of poses.

The goal of \acrshort{PGO} is to find a solution $\mathbf{x}^*$ that minimizes the following objective function:
\begin{equation}
    \label{eq:full_problem}
    \mathbf{x}^* = \argmin\sum_{(i,j) \in \mathcal{E}} \mathbf{e}_{i,j}(\mathbf{x}_i, \mathbf{x}_j)^T \mathbf{\Omega}_{i,j}  \mathbf{e}_{i,j}(\mathbf{x}_i, \mathbf{x}_j),
\end{equation}
where $\mathbf{\Omega}_{i,j}$ denotes the information matrix representing the uncertainty associated with the measurement $\mathbf{z}_{i,j}$, which correlates the state variables $\mathbf{x}_i$ and $\mathbf{x}_j$. The corresponding error is defined as follows: 
\begin{equation}
    \label{eq:general_error}
   \mathbf{e}_{i,j}(\mathbf{x}_i, \mathbf{x}_j) = \mathbf{h}(\mathbf{x}_i, \mathbf{x}_j) \boxminus \mathbf{z}_{i,j},
\end{equation}
where $\mathbf{h}(\mathbf{x}_i, \mathbf{x}_j)$ represents the measurement function, which computes the expected measurement based on the current state estimates. In the context of \acrshort{PGO}, $\mathbf{h}(\mathbf{x}_i, \mathbf{x}_j)$ computes the rigid transformation between the two poses $\mathbf{x}_i$ and $\mathbf{x}_j$. The term $\mathbf{z}_{i,j}$ denotes the odometry or loop closure measurement provided by the front-end module, and $\boxminus$ indicates the difference operator defined over the underlying manifold.

The solution to \eqref{eq:full_problem} is obtained by solving a \acrlong{NLLS} (\acrshort{NLLS}) problem using minimization algorithms such as Gauss-Newton~\cite{burke1995gauss}, Levenberg-Marquardt~\cite{more2006levenberg}, or Powell’s Dog Leg~\cite{nocedal1999numerical}.
Throughout the text, $\mathbf{x}_i$ denotes both a robot pose and its corresponding graph node. Similarly, $\mathbf{e}_{i,j}(\mathbf{x}_i, \mathbf{x}_j)$ denotes the error term, measurement, and constraint, as well as the edge $(i,j) \in \mathcal{E}$.

Following~\cite{latif2013robust}, we partition the edges in two sets, $\mathcal{E} = \mathcal{E}_o \cup \mathcal{E}_l $ with $ \mathcal{E}_o \cap \mathcal{E}_l = \emptyset $, where $\mathcal{E}_o = \{ (i,j) \in \mathcal{E} \wedge i < j \mid j = i + 1\}$ is the set of edges associated with the odometry measurements, and $\mathcal{E}_l = \{ (i,j) \in \mathcal{E} \wedge i < j \mid j \neq i + 1\}$ is the set of edges associated with the loop closure measurements. 
To make the notation more fluid the following abbreviations are used: 
\begin{itemize}[noitemsep,topsep=0pt,after=\vspace{5pt}]
    \item $\mathbf{e}_{i, i+1} = \mathbf{e}_{i, i+1}(\mathbf{x}_i, \mathbf{x}_{i+1})$ for the odometry errors;
    \item $ \mathbf{e}_{i,j} = \mathbf{e}_{i,j}(\mathbf{x}_i, \mathbf{x}_{j})$ for the loop closures.
\end{itemize} 
We can rewrite \eqref{eq:full_problem} as:
\begin{equation}    
    \label{eq:offline_division1}
    \begin{gathered}
    \mathbf{x}^* = \argmin\sum_{(i,i+1) \in \mathcal{E}_o} \mathbf{e}_{i,i+1}^T \mathbf{\Omega}_{i,i+1}  \mathbf{e}_{i,i+1} \\ + \sum_{(i,j) \in {\mathcal{E}}_l} \mathbf{e}_{i,j}^T \mathbf{\Omega}_{i,j}  \mathbf{e}_{i,j},
    \end{gathered}
\end{equation}
where in an ideal case only \emph{inlier} loop closure measurements should be taken into account. The goal of our approach is to minimize the probability of integrating loop closure outliers during optimization.
\section{The \method Framework: Test and Check for Robust PGO}
\label{sec:methodology}
The \acrlong{TACO} (\acrshort{TACO}) framework (\figref{fig:teaserTACO}) belongs to the class of consistent set maximization methods for robust pose graph optimization (see Section~\ref{subsec:maximalset}). \acrshort{TACO} is composed of two main components that work in synergy. 

Section~\ref{subsec:IPC} introduces the \textit{test} component, namely the \acrlong{IPC} (\acrshort{IPC}) algorithm, evaluates the consistency of each incoming loop closure in an online manner, incrementally building an approximation of the maximal set of consistent measurements.

Section~\ref{sec:so-sc} then presents the \textit{check} component, namely the \acrlong{SOS} (\acrshort{SOS}), leverages the existing Switchable Constraints~\cite{sunderhauf2012switchable} to revise the last $M$ loop closures accepted by \acrshort{IPC}, correcting potentially degenerate estimates that would impact subsequent \acrshort{IPC} decisions. As a result, the set of consistent measurements produced by \acrshort{IPC} is further refined by removing residual outliers.

Before presenting the details of \acrshort{IPC} and \acrshort{SOS}, we first introduce the assumptions and heuristics from robust optimization literature that guided the design of the proposed framework. These are subsequently referenced where they are applied within \acrshort{IPC} and \acrshort{SOS}.

\subsection{Assumptions and Heuristics}
\label{sec:assumption}
The \acrshort{TACO} framework is built on the following assumptions:

\textbf{A1}: \textit{Odometry measurements are inliers.} Odometry measurements are affected by different levels of Gaussian noise, whose magnitude depends on the quality of the front-end. However, as noted in~\cite{carlone2014selecting}, they are free of outliers, thus greatly reducing the complexity of assessing measurement consistency. Furthermore, these measurements can be leveraged to enforce maintaining the local shape of the trajectory.

\textbf{A2}:\textit{ Prevalence of inliers.} 
Given the inherent aliasing effect in the environment, it is natural for a place recognition module to produce a certain percentage of outliers. On the other hand, it is reasonable to expect that the majority of loop closure measurements are inliers.
Furthermore, considering \textbf{A1} and the higher frequency of odometry measurements relative to loop closure measurements, the overall prevalence of inlier measurements is ensured.

In \acrshort{TACO}, we incorporate the following heuristics drawn from the existing literature on robust pose graph optimization:

\textbf{H1}:\textit{ Start from a simpler version of the problem, for which its solution will be meaningful for the more complex version of the problem.} As stated in~\cite{grisetti2012robust, guadagnino2021hipe}, a simpler version of a problem exhibits a broader and more stable convergence basin, thus ensuring better convergence. The solution obtained for the reduced problem can be used as the initial guess for the full problem, thus improving the convergence of the latter.

\textbf{H2}:\textit{ The nature of the compatibility test is to reject as many outliers as possible while preserving the inliers.} Since the distinction between outliers and inliers is not always observable~\cite{carlone2014selecting}, consistency between measurements is evaluated instead of explicitly detecting outliers.

\subsection{The \texorpdfstring{\acrlong{IPC}}{Incremental Probabilistic Consensus} Algorithm}
\label{subsec:IPC}
The \acrlong{IPC} (\acrshort{IPC}) algorithm constitutes the \textit{test} component of \acrshort{TACO}. Its purpose is to assess the consistency of incoming loop closures in an online fashion, allowing for rapid integration of measurements and timely updates of the refined pose estimate.\\
To evaluate consistency, \acrshort{IPC} leverages the set of previously accepted loop closure measurements, referred to as the \textbf{consensus set} $\mathcal{C}$.

The following steps outline the \acrshort{IPC} algorithm. Each step reports, in square brackets, the corresponding assumptions and heuristics (see \secref{sec:assumption}) and the relevant pseudocode lines in \algref{alg:cap}:
\begin{itemize}
    \item \textbf{[Lines~\ref{alg:odometry}-\ref{alg:update_Eo}]}: Integrates all odometry measurements and generates a new node when necessary (for example, if the distance to the previous node is above a certain threshold) [\textbf{A1}];
    \item \textbf{[Lines~\ref{alg:loop}-\ref{alg:end_aff}]}: For each incoming loop closure measurement $\textbf{z}_{i,j}$, compute the (minimal) independent subgraph (see \secref{sec:issp}) that contains the relative odometry constraints and loop closure constraint in $\mathcal{C}$ collected so far; 
    \item \textbf{[Line~\ref{alg:opt}]}: Solve the \acrshort{PGO} problem only for the independent subgraph [\textbf{H1}];
    \item \textbf{[Lines~\ref{alg:consensus}-\ref{alg:full_update}]}: Accept and include $\textbf{z}_{i,j}$ in $\mathcal{C}$ only if \emph{all} measurements in the independent subgraph are consistent (\ie \emph{agree}, see below) with the solution obtained in the previous point [\textbf{A1}, \textbf{A2}, \textbf{H1}, \textbf{H2}];
    \item \textbf{[Line~\ref{alg:propagate}]}: If $\textbf{z}_{i,j}$ is accepted, propagate the obtained solution to the remaining nodes as described in \secref{sec:issp}. 
    \item \textbf{[Lines~\ref{alg:else_restore}-\ref{alg:restore}]}: If $\textbf{z}_{i,j}$ is rejected, restore the nodes in the independent subgraph to their state prior to the optimization.
\end{itemize}

\begin{algorithm}[t!]
\DontPrintSemicolon
\caption{IPC}\label{alg:cap}
\small
$\mathcal{C}, \mathcal{E}_o, \mathcal{E}_l \leftarrow \emptyset$\;
$\mathcal{V} \leftarrow \{\mathbf{x}_0\}$\;
\Upon{receiving $\mathbf{z}_{i,j}$}
{
    \tcp{odometry measurement}
    \If{ $j = i + 1$  }
    {\label{alg:odometry}
        $\mathbf{x}_{i+1} \leftarrow \mathbf{x}_i \boxplus \mathbf{z}_{i,i+1}$ \tcp{integrate motion} \label{alg:motion}
        $\mathbf{e}_i \leftarrow \texttt{edge}(\mathbf{x}_{i}, \mathbf{x}_{i+1}, \mathbf{z}_{i,i+1}, \mathbf{\Omega}_{i,i+1})$\;
        $\mathcal{E}_o \leftarrow \mathcal{E}_o \cup \{\mathbf{e}_i\}$ \tcp{update edges} 
        $\mathcal{V} \leftarrow \mathcal{V} \cup \{\mathbf{x}_{i+1}\}$ \tcp{update nodes} \label{alg:update_Eo}
    }
    \tcp{loop closure measurement}
    \Else
    {   \label{alg:loop}
        \tcp{$\mathcal{G}^I$ initialized to $\mathcal{G}(i,j)$}
        $\mathbf{e}_{i,j} \leftarrow \texttt{edge}(\mathbf{x}_{i}, \mathbf{x}_{j}, \mathbf{z}_{i,j}, \mathbf{\Omega}_{i,j}) $\;
        $\mathcal{V}^I, \mathcal{E}^I, \mathcal{G}^I \leftarrow \mathcal{G}(i,j)$\; \label{alg:initAff}
        $ a \leftarrow i $\;
        \tcp{add edges until there exists an}
        \tcp{edge that has a node in both}
        \tcp{$\mathcal{V}^I$ and $\mathcal{V} \setminus \mathcal{V}^I$}
        \While{$\exists \mathbf{e}_{k,g} \in \mathcal{C} \setminus \mathcal{E}^I(a,b) \mid \mathbf{x}_k \in \mathcal{V} \setminus \mathcal{V}^I(a,b) \wedge \mathbf{x}_g \in \mathcal{V}^I(a+1,b)$}
        {\label{alg:independence}
            $\mathcal{E}^I \leftarrow \mathcal{E}^I \cup \{\mathbf{e}_k,\dots, \mathbf{e}_a\}$\;
            $\mathcal{E}^I \leftarrow \mathcal{E}^I \cup \{\mathbf{e}_{k,g}\}$\;
            $\mathcal{V}^I \leftarrow \mathcal{V}^I \cup \{\mathbf{x}_k,\dots, \mathbf{x}_{a-1}\}$\;
            $ a \leftarrow k$
        }\label{alg:end_aff}
        \texttt{solve} \eqref{eq:complex_consensus} \tcp{PGO on $\mathcal{G}^I$} \label{alg:opt} 
        \If{ $\mathbf{e}_{a,b}^T \mathbf{\Omega}_{a,b} \mathbf{e}_{a,b} < \chi^2_{\alpha, \delta}, \forall \mathbf{e}_{a,b} \in \mathcal{E}^I$ } 
        {\label{alg:consensus}
            $\mathcal{C} \leftarrow \mathcal{C} \cup \{\mathbf{e}_{ij}\}$\; \label{alg:consSet}
            $\mathcal{E}_l \leftarrow \mathcal{E}_l \cup \{\mathbf{e}_{ij}\}$\; \label{alg:full_update}
            propagate solution to $\mathbf{x}_k \in \mathcal{V} \setminus \mathcal{V}^I$ as described in \secref{sec:issp} \label{alg:propagate}
        }
        \Else
        {\label{alg:else_restore}
            $\texttt{restore}(\mathcal{V}^I)$\; \label{alg:restore}
        }
    }
}
\end{algorithm}

\subsubsection{Independent Subgraphs\texorpdfstring{\\}{}}
\label{sec:issp}
Since the pose graph encodes the underlying optimization problem in graphical form, solving one of its partitions is equivalent to addressing a reduced subproblem of the full formulation. One of the foundational components of \acrshort{IPC} is the identification of a minimal yet informative partition of the pose graph for assessing the consistency of incoming loop closure constraints in an online manner. Furthermore, the solution obtained on this partition can be propagated and serves as an effective initialization for solving the full problem [\textbf{H1}]. This section describes how such partitions are constructed.

\begin{figure}[!t]
    \centering
    \includegraphics[width=0.5\textwidth]{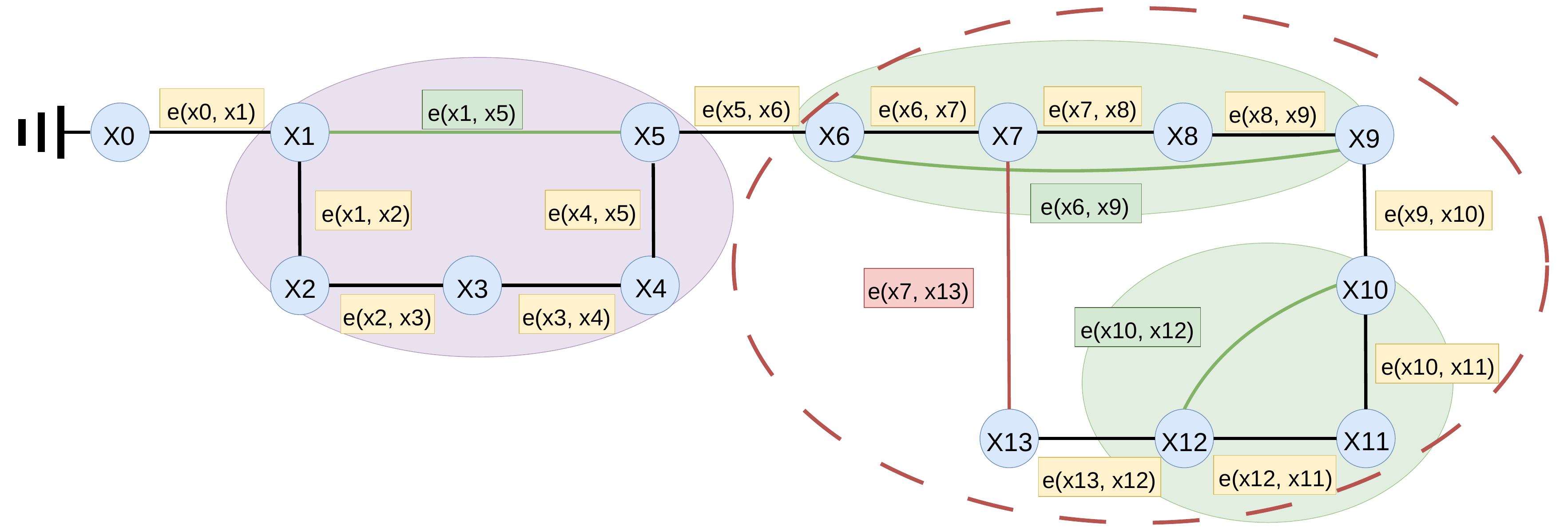}
    \caption{An example of a pose graph with nodes represented by light blue circles. Each error term is associated with the corresponding edge. The coordinate system is usually fixed in the first node $\mathbf{x}_0$. Light green and purple ovals represent examples of simple independent subgraphs (\ie they only include one loop) while the dotted red oval represents more complex independent subgraphs that include both multiple loops with crossing edges and internal loops.}
    \label{fig:strict_consensus}
\end{figure}

We define a subgraph $\mathcal{G}(a,b) = (\mathcal{V}(a,b),\mathcal{E}(a,b))$ where $\mathcal{V}(a,b)=\{i \in \mathcal{V} \mid a \leq i \leq b\}$ and $\mathcal{E}(a,b)=\{ (i,j) \in \mathcal{E} \mid a \leq i < j \leq b\}$. By construction, such a subgraph defines a time interval that includes all nodes, and related edges, from index $a$ to index $b$, with $a<b~\text{and}~b-a > 1$.\\
A subgraph $\mathcal{G}^I(a,b)$ is \emph{independent} if the following condition is satisfied:
\begin{equation}
\label{eq:independence}
    \nexists \mathbf{e}_{i,j} \in \mathcal{E} \setminus \mathcal{E}^I(a, b) \mid \mathbf{x}_i \in \mathcal{V} \setminus \mathcal{V}^I(a,b) \wedge \mathbf{x}_j \in \mathcal{V}^I(a+1,b).
\end{equation}
The condition in \eqref{eq:independence} enforces that $\mathcal{G}^I(a,b)$ is connected to its complementary partition only at its endpoints through odometry edges. As a result, optimization over $\mathcal{G}^I(a,b)$ can be treated as a self-contained subproblem for measurement consistency evaluation (see \figref{fig:strict_consensus}), and its solution can be \emph{propagated} to better initialize the remaining nodes via forward odometry composition from $\mathbf{x}_b$.

To construct the independent subgraph $\mathcal{G}^I(a,b)$ corresponding to the loop closure edge $\mathbf{e}_{i,j} \in \mathcal{E}_l$, we first initialize $\mathcal{G}^I(a,b)$ with $a=i$ and $b=j$. The subgraph is then iteratively expanded by incorporating edges $\mathbf{e}_{k,g} \in \mathcal{E} \setminus \mathcal{E}^I(a,b)$, together with their corresponding nodes $\mathcal{V}(k,g) \setminus \mathcal{V}^I(a, b)$, whenever they violate the independence condition in \eqref{eq:independence}. This process continues until no violating edges remain, ensuring that $\mathcal{G}^I(a,b)$ satisfies the independence property.

Within \acrshort{IPC}, this procedure is significantly accelerated (lines~\ref{alg:loop}-\ref{alg:end_aff} of \algref{alg:cap}) by restricting the independence verification to loop closures in the consensus set $\mathcal{C}$ that are not already included in $\mathcal{E}^I(a,b)$ (line~\ref{alg:independence}). This simplification is justified as, once initialized, odometry edges cannot violate the independence condition.

In the example in \figref{fig:strict_consensus}, for loop closure $\mathbf{e}_{7, 13}$, the independent subgraph is initialized to $\mathcal{G}^I(7,13)$. Since $\mathbf{e}_{7,13}$ violates the independence condition, the subgraph is expanded to $\mathcal{G}^I(6,13)$. The resulting subgraph satisfies the independence property, as both $\mathcal{G}^I(6,13)$ and its complement $\mathcal{G} \setminus \mathcal{G}^I(6,13)$ are connected only through the odometry edge $\mathbf{e}_{5,6}$.

\subsubsection{Measurement Consistency Test\texorpdfstring{\\}{}}
Another foundational idea behind \acrshort{IPC} is the concept of \textit{consensus}. In the absence of outliers and assuming a sufficiently accurate initial estimate, the solution of the \acrshort{PGO} problem yields a set of poses $[\mathbf{x}_1,\dots,\mathbf{x}_n]$ for which the residual errors associated with all edges in $\mathcal{E}$ are consistent with their Gaussian noise models~\cite{5681215}.
In this setting, the measurement set $\mathcal{E}$ is said to have reached consensus, or to agree, with \acrshort{PGO} solution of $\mathcal{G}$. 
More formally, $\mathcal{E}$ is said to reach consensus over the \acrshort{PGO} solution of $\mathcal{G}$ if and only if each edge in $\mathcal{G}$ satisfies the $\chi^2$ test~\cite{neira2001data}:
\begin{equation}
    \label{eq:consensus}
    \mathbf{e}_{i,j}^T \mathbf{\Omega}_{i,j} \mathbf{e}_{i,j} < \chi^2_{\alpha, \delta}, \forall (i,j) \in \mathcal{E}
\end{equation}
where $\alpha$ is the confidence parameter and $\delta$ the degrees of freedom of the error.

Prior to incorporating loop closures, \acrshort{IPC} operates on an edge set that is inherently in consensus, since pose estimates are obtained through odometry integration and odometry constraints are assumed to be outlier-free [\textbf{A1}]. Consequently, \acrshort{IPC}'s objective is to preserve this state of consensus throughout the robot's navigation, while incrementally building the set $\mathcal{C}$, referred to as the consensus set.

Since odometry constraints are inherently consistent [\textbf{A1}], consistency verification focuses exclusively on loop closure edges. A loop closure $\mathbf{e}_{i, j}$ is included in $\mathcal{C}$ if and only if the \acrshort{PGO} solution over the associated independent subgraph $\mathcal{G}^I(a,b)$ satisfies \eqref{eq:consensus}.

After identifying $\mathcal{G}^I(a, b)$, its edge set is partitioned as $\mathcal{E}^I(a, b) = \mathcal{E}_o^I(a, b) \cup \mathcal{E}_l^I(a, b)$, separating odometry and loop closure constraints as done in~\eqref{eq:offline_division1}. Most existing methods~\cite{yang2020graduated, sunderhauf2012switchable, ramezani2022aeros, agarwal2013robust, olson2013inference} mitigate outliers by introducing robust cost functions on loop closures. Robust cost functions introduce additional non-linearities and require estimating individual re-weighting factors for each loop closure, thus increasing computational complexity and sensitivity to initialization. 

Instead, \acrshort{IPC} strengthens the contribution of the \textit{trustworthy} (i.e.~more reliable) odometry measurements through a fixed scaling factor $s > 1$ applied to their information matrices:
\begin{equation}
        \label{eq:complex_consensus}
    \sum_{(i, i+1) \in \mathcal{E}_o^I(a, b)} \mathbf{e}_{i, i+1}^T s\mathbf{\Omega}_{i,i+1}  \mathbf{e}_{i, i+1} +
    \sum_{(i,j) \in \mathcal{E}_l^I(a, b)} \mathbf{e}_{i,j}^T \mathbf{\Omega}_{i,j}  \mathbf{e}_{i,j}.
\end{equation}
This factor is shared across all odometry edges and does not introduce additional non-linearities, preserving the original problem’s complexity. The factor $s$ is used only to guide the optimization process, while consistency is evaluated using the original measurement noise model.

This formulation can also be interpreted as applying a scaled $L_2$ M-estimator to the odometry errors, namely: 
\begin{equation}
        \label{eq:ipc_mestimator}
    \rho_{IPC}(e(x)) = s\frac{e(x)^2}{2}.
\end{equation}
The corresponding weight function $\gamma(x)_{IPC}$, used to re-scale the information matrix as $\hat{\mathbf{\Omega}}_{i,i+1} = \gamma(x)_{IPC}\mathbf{\Omega}_{i,i+1}$, is therefore given by: 
\begin{equation}
    \label{eq:gamma_factor}
    \gamma(x)_{IPC} = \frac{1}{e(x)}\frac{ \partial p(e)}{ \partial e}\big{|}_{e=e(x)}  =  s.
\end{equation}

\subsubsection{IPC's Limitations\texorpdfstring{\\}{}}
The primary limitation of \acrshort{IPC} stems from the fact that accepted loop closures are never re-evaluated. If an outlier is mistakenly included in $\mathcal{C}$, future consistency checks may be negatively affected. This risk is highest during the early stages of incremental loop closure validation, when the inlier set $\mathcal{C}$ contains few or no elements. 

In such cases, the minimal independent subgraph $\mathcal{G}^I(a, b)$ used for the consistency check may consist only of odometry edges. Depending on $||\mathcal{E}_o^I(a, b)||$, the odometry edges may not impose enough constraints to prevent the integration of outliers. The relation between the cardinality of $\mathcal{E}_o^I(a, b)$ and the likelihood of integrating an outlier follows from:
\begin{equation}    
    \label{eq:odom_only_opt}
    \begin{gathered}
    F(\mathbf{x}) = \sum_{(i,i+1) \in \mathcal{E}^I_o(a, b)} \mathbf{e}_{i,i+1}^T \mathbf{\Omega}_{i,i+1}  \mathbf{e}_{i,i+1} + \mathbf{e}_{i,j}^T \mathbf{\Omega}_{i,j}  \mathbf{e}_{i,j}.
    \end{gathered}
\end{equation}
Satisfying the consistency condition \eqref{eq:consensus} translates to finding any node configuration $\mathbf{x}^*$ such that:
\begin{equation}    
    \label{eq:odom_only_sol}
    \begin{gathered}
    F(\mathbf{x}^*) \leq (||\mathcal{E}_o^I(a, b)|| + 1)\chi^2_{\alpha, \delta}.
    \end{gathered}
\end{equation}
This expression defines the maximum admissible error that an outlier can introduce without violating the consistency test. 

Since in this scenario all state variables in $\mathcal{V}^I(a, b)$ are initialized purely from the odometry measurements, the first term of \eqref{eq:odom_only_opt} is zero at the start of optimization.

As the \acrshort{NLLS} solver seeks a solution that satisfies all constraints, the resulting error may be distributed over the entire odometry chain, allowing a spurious loop closure to pass the test. Consequently, \acrshort{IPC} becomes vulnerable when operating over long, open-loop trajectories.
\subsection{SOS: Switchable Outlier Sanitization} 
\label{sec:so-sc}
To overcome \acrshort{IPC}'s limitations, we propose integrating a \textit{check} component capable of retrospectively revising the decisions made on previously evaluated loop closures. This component dubbed \acrlong{SOS} (\acrshort{SOS}) leverages the existing \emph{Switchable Constraints (SC)} to evaluate measurement consistency for the purpose of sanitizing any outliers mistakenly included in $\mathcal{C}$. 

\acrshort{SOS} runs periodically on a set we call the \textit{unrevised set} ${\mathcal{U}_r}$ (see \figref{fig:teaserTACO}), which contains the last $M$ loop closures evaluated \textbf{as consistent} by \acrshort{IPC} \textit{but not yet revised with} \acrshort{SOS}. On the other hand, the \emph{revised set} $\mathcal{R}_c$ contains all the inlier loop closures that were \textit{revised} at previous iterations using \acrshort{SOS}. Therefore, $\mathcal{C} = \mathcal{R}_c \cup \mathcal{U}_r$, with $\mathcal{R}_c \cap \mathcal{U}_r = \emptyset$.\\
Similarly to \acrshort{IPC}, \acrshort{SOS} uses $\mathcal{R}_c$ to identify a meaningful subgraph for evaluating measurement consistency [\textbf{H1}].

The following steps outline the \acrshort{SOS} algorithm, with each step reporting, in square brackets, the corresponding assumptions and heuristics (see \secref{sec:assumption}) and the relevant pseudocode lines in \algref{alg:sos}:
\begin{itemize}
    \item \textbf{[Lines~\ref{alg:mandatory_vertices_init}-\ref{alg:find-subg}]}: Compute the (minimal) trusted subgraph identified by the loop closure edges in $\mathcal{U}_r$ (see \secref{sec:trusted_sub}) [\textbf{H1}];
    \item \textbf{[Lines~\ref{alg:enhance_init}-\ref{alg:solve_enhanced}]}: Replace the loop closure edges in $\mathcal{U}_r$ with their respective switchable variants and solve the augmented \acrshort{PGO} problem on the trusted subgraph (see \secref{sec:switchable});
    \item \textbf{[Lines~\ref{alg:for2-start}-\ref{alg:empty_set}]}: Reject from the consensus set $\mathcal{C}$ all loop closure measurements that are not mutually consistent (see \secref{sec:outlier_sanity}) [\textbf{H2}].
\end{itemize}

\subsubsection{Switchable Constraints\texorpdfstring{\\}{}}
\label{sec:switchable}
\begin{figure}[t]
    \centering
    \includegraphics[width=0.45\textwidth]{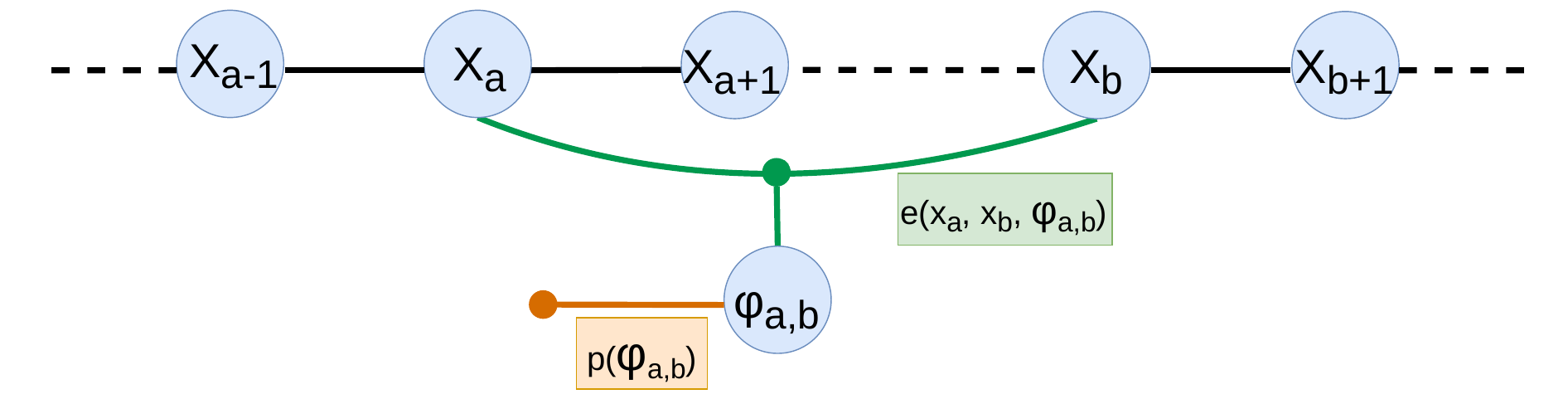}
    \caption{This figure illustrates the switchable constraint formulation described in~\cite{sunderhauf2012switchable}. The green edge represents the ternary factor involving the variables $\mathbf{x}_a$, $\mathbf{x}_b$, and $\varphi_{a,b}$, where $\varphi_{a,b}$ serves as the switch variable, capable of activating or deactivating the constraint. The orange unary edge connected to $\varphi_{a,b}$ represents the prior constraint on the loop closure edge.}
    \label{fig:switchable_constraint}
\end{figure}

\acrshort{SOS} builds upon the \emph{Switchable Constraints (SC)} framework introduced in~\cite{sunderhauf2012switchable}. SC augments the standard pose graph optimization problem by enabling the graph topology to change during optimization. This is achieved with the introduction of a new set of variables and constraints: 
\begin{equation}
    \label{eq:swc_constraints_1}
     \hat{\mathbf{e}}_{i,j} = \mathbf{e}(\mathbf{x}_i, \mathbf{x}_j, \varphi_{i,j}) = \Psi(\varphi_{i,j})\mathbf{e}_{i,j}(\mathbf{x}_i, \mathbf{x}_j),
\end{equation}
\eqref{eq:swc_constraints_1} is an extended version of the loop closure error factor in \eqref{eq:offline_division1}. While the original loop closure factor depends only on the variables $\mathbf{x}_i$ and $\mathbf{x}_j$, \eqref{eq:swc_constraints_1} introduces an additional variable, $\varphi_{i,j} \in \mathbb{R}$,  referred to as the \textit{switch} variable. This variable modulates the weight of the loop closure constraints through the function $\Psi(\cdot): \mathbb{R} \rightarrow [0, 1]$, which remaps the switch variable into the $[0, 1]$ interval. Ideally, $\Psi(\varphi_{i,j}) = 1$ when the loop closure $\mathbf{z}_{i,j}$ is an inlier, and $\Psi(\varphi_{i,j}) = 0$ otherwise. Setting $\Psi(\varphi_{i,j}) = 0.0$ is equivalent to removing the topological connection between $\mathbf{x}_{i}$ and $\mathbf{x}_{j}$ introduced by $\mathbf{z}_{i,j}$ (see \figref{fig:switchable_constraint}).

As noted in~\cite{sunderhauf2012switchable}, better convergence can be obtained by selecting the simple linear function $\Psi(\varphi_{i,j}) = \varphi_{i,j}$, rather than sigmoid or step functions, as long as we constrain $0 \leq \varphi_{i,j} \leq 1$.
To avoid the trivial solution where all switch variables collapse to zero, an additional penalty term is introduced in \eqref{eq:offline_division1}:
\begin{equation}
    \label{eq:swc_constraints_2}
     p_{i,j} = p(\varphi_{i,j}) = || \eta_{i,j} - \varphi_{i,j} ||.
\end{equation}
This penalization term relies on $\eta_{i,j} \in [0,1]$, which encodes prior belief on the inlier or outlier nature of the loop constraint $\mathbf{e}_{i,j}$.
Finally, the full augmented problem is obtained by replacing the second error term in \eqref{eq:offline_division1} with both \eqref{eq:swc_constraints_1} and \eqref{eq:swc_constraints_2}: 
\begin{equation}
    \label{eq:full_switchable}
    \begin{gathered}
   \sum_{(i,i+1) \in \mathcal{E}_o} \mathbf{e}_{i, i+1}^T\mathbf{\Omega}_{i, i+1} \mathbf{e}_{i, i+1} \; +\\ \sum_{(i,j) \in \mathcal{E}_l} ( \hat{\mathbf{e}}_{i,j}^T \mathbf{\Omega}_{i,j}  \hat{\mathbf{e}}_{i,j} + p_{i,j}\Omega_{p_{i,j}} p_{i,j} ).
   \end{gathered}
\end{equation}
Direct integration of SC poses two major challenges. First, it increases the number of variables to estimate in an already highly non-convex SLAM problem. This added complexity results in a more irregular convergence basin~\cite{grisetti2012robust}, making it harder for the optimizer to reach a favorable local optimum. Second, the default prior value $\eta_{i,j}$ of $\varphi_{i,j}$, is set to 1. This biases the optimizer toward accepting as many loop closures as possible. In environments with a high percentage of outliers, such as caves or underwater environments, this strategy for setting priors becomes suboptimal.

\begin{algorithm}[t!]
\DontPrintSemicolon
\caption{SOS($\mathcal{G}$, $\mathcal{R}_c$, $\mathcal{U}_r$)}
\label{alg:sos}
\small
\tcp{Set of mandatory vertices to visit}
$\mathcal{V}_r \leftarrow \emptyset$ \; \label{alg:mandatory_vertices_init}
$a = \texttt{inf}, b = 0$ \;
\For {( each $\mathbf{e}_{i,j} \in \mathcal{U}_r$ )}
{
    $a = \texttt{min}(a, i)$ \;
    $b = \texttt{max}(b, j)$ \;
    $\mathcal{V}_r \leftarrow \mathcal{V}_r \cup \{ \mathbf{x}_i, \mathbf{x}_j \}$ \;
}
\tcp{Compute trusted subgraph}
$\mathcal{G}^T \leftarrow \mathcal{G}(a, b)$ \;
$\mathcal{G}^T \leftarrow \textsc{PruneGraphDijkstra}(\mathcal{G}^T, \mathcal{V}_r, \mathcal{R}_c)$\; \label{alg:find-subg}
\tcp{Augment $\mathcal{G}^T$ with SC}
\For {( each $\mathbf{e}_{i,j} \in \mathcal{U}_r$ )} 
{\label{alg:enhance_init}
    $\mathcal{V}^T \leftarrow \mathcal{V}^T \cup \{ \varphi_{i,j} \}$  \label{alg:switch_variable} \;
    $\mathcal{E}^T \leftarrow \mathcal{E}^T \cup \{ \hat{\mathbf{e}}_{i,j}, p_{i,j} \}$  \label{alg:switch_prior} \;
}
solve \eqref{eq:subgraph-srsc} \; \label{alg:solve_enhanced}
\tcp{Discard inconsistent loop closures}
\For {( each $\mathbf{e}_{i,j} \in \mathcal{U}_r$ )}
{\label{alg:for2-start}
  \If{ $\varphi_{i,j} \geq th_c $  }
  {
    $\mathcal{R}_c \leftarrow \mathcal{R}_c \cup \{\mathbf{e}_{i,j} \}$ \;\label{alg:for2-end}
  }
}
\tcp{Clear the unrevised set}
$ \mathcal{U}_r \leftarrow \emptyset$ \;\label{alg:empty_set}
\end{algorithm}

\begin{figure*}[!t]
    \centering
    \includegraphics[width=0.8\linewidth]{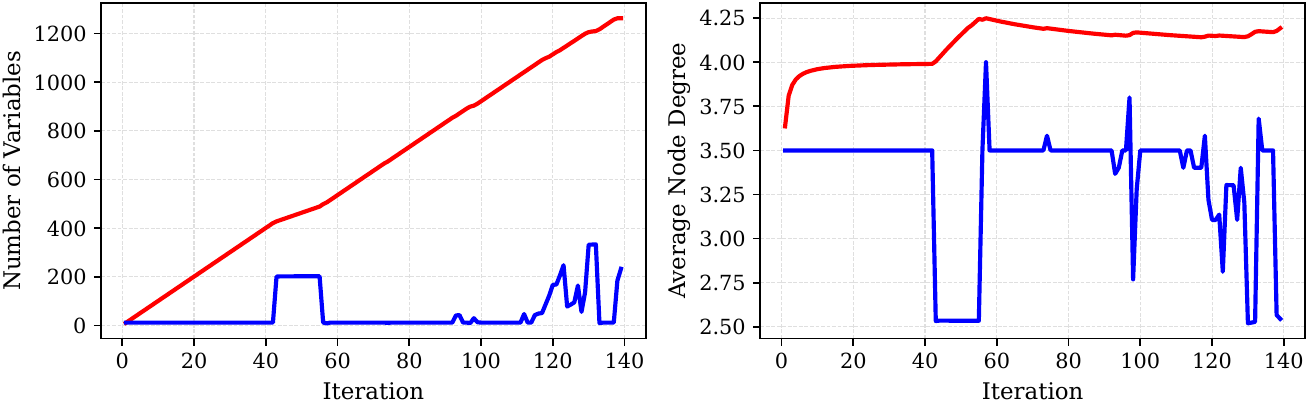}
    \caption{The left plot compares the number of variables of the full problem in$\mathcal{G}$ (red) and the trusted subgraph $\mathcal{G}^T$ (blue) across \acrshort{SOS} iterations, while the right plot reports the corresponding average node degree. Results are obtained on the FRH dataset with 50\% injected outliers.
    }
    \label{fig:node_degree}
\end{figure*}

\subsubsection{Trusted Subgraphs\texorpdfstring{\\}{}}
\label{sec:trusted_sub}
To address the increased complexity of the \acrshort{PGO} problem introduced by SC variables, \acrshort{SOS}, similarly to \acrshort{IPC}, identifies a minimal subgraph for evaluating the consistency of loop closure edges in the unrevised set $\mathcal{U}_r$ [\textbf{H1}]. 
The goal is to obtain a connected subgraph with the fewest possible nodes, which can be interpreted as an instance of the \textit{Steiner problem}~\cite{salazar2003steiner}.  

The subgraph must include all edges in $\mathcal{U}_r$ and their associated nodes $\mathcal{V}_r = \{ i \in \mathcal{V} \mid \exists \mathbf{e}_{k,g} \in \mathcal{U}_r: k=i \vee g=i \}$ as their consistency is evaluated. To guarantee the existence of a solution of \acrshort{PGO} problem, its corresponding graph must be connected~\cite{carlone2014fast}. However, edges in the unrevised set $\mathcal{U}_r$ are replaced by their SC variants, which may deactivate connections and disconnect portions of the subgraph (lines~\ref{alg:find-subg}-\ref{alg:enhance_init} of \algref{alg:sos}). To address this, the subgraph connectivity must be guaranteed to use only odometry edges and revised loop closure edges $\mathcal{E}_o \cup \mathcal{R}_c$, which do not rely on SC. Under this condition, we define a \textit{trusted} subgraph $\mathcal{G}^T$ as subgraph that remains connected when the loop closure edges in $\mathcal{U}_r$ are not considered.

The simplest trusted subgraph is $\mathcal{G}^T = \mathcal{G}(a,b)$, defined in \secref{sec:issp}, where $a = \min(\mathcal{V}_r)$ and $b = \max(\mathcal{V}_r)$. Connectivity is guaranteed by the odometry chain between these nodes. $\mathcal{G}^T = \mathcal{G}(a,b)$ defines the base graph for consistency evaluation and typically involves far fewer nodes than $\mathcal{G}$, \ie $||\mathcal{V}^T|| << ||\mathcal{V}||$. Starting from this set, Dijkstra’s algorithm~\cite{dijkstra2022note} is used to further prune unnecessary nodes and odometry edges by exploiting trusted loop edges in $\mathcal{R}_c$ as shortcuts (lines~\ref{alg:mandatory_vertices_init}-\ref{alg:find-subg}).

As illustrated in \figref{fig:node_degree}, the average node degree of $\mathcal{G}^T$ is comparable to that of the full graph. Since the average node degree is directly proportional to the likelihood of estimating a globally optimal \acrshort{PGO} solution~\cite{olson2009evaluating, khosoussi2019reliable}, $\mathcal{G}^T$ is expected to have a similar probability of reaching the global optimum as the full graph $\mathcal{G}$. Furthermore, reducing both the number of switchable variables and the size of the optimization problem leads to faster optimization and a higher likelihood of convergence~\cite{grisetti2012robust, guadagnino2021hipe}.

\subsubsection{Outlier Sanitization\texorpdfstring{\\}{}}
\label{sec:outlier_sanity}
Instead of using SC for inlier-outlier detection as in~\cite{sunderhauf2012switchable}, in \acrshort{SOS} it is leveraged to sanitize outliers mistakenly introduced by \acrshort{IPC}. In this setting, the strategy of setting the prior $\eta_{i,j} = 1$ is effective, as SC is operating on a pre-filtered set of loop closures already deemed consistent by \acrshort{IPC}.
To assess the consistency of loop edges in $\mathcal{U}_r$ within the trusted subgraph, we reformulate the objective in \eqref{eq:full_switchable} as
\begin{equation}
    \label{eq:subgraph-srsc}
    \begin{gathered}
   \sum_{(i,j) \in \mathcal{E}^T \setminus \mathcal{U}_r } \mathbf{e}_{i,j}^T \mathbf{\Omega}_{i,j}  \mathbf{e}_{i,j}
    +\\
   \sum_{(i,j) \in \mathcal{U}_r} ( \hat{\mathbf{e}}_{i,j}^T \mathbf{\Omega}_{i,j}  \hat{\mathbf{e}}_{i,j} + p_{i,j}\Omega_{p_{i,j}} p_{i,j} ).
   \end{gathered}
\end{equation}
The switch variables $\varphi_{i,j}$ are interpreted as consistency scores, quantifying the agreement of each edge $\mathbf{e}_{i,j}$ with the rest of the edges in $\mathcal{E}^T$. After optimization, edges $\mathbf{e}_{i,j} \in \mathcal{U}_r$ with scores below $th_c = 0.9$ are discarded and excluded from $\mathcal{C}$ (lines~\ref{alg:solve_enhanced}-\ref{alg:empty_set} of \algref{alg:sos}). If any edge in $\mathcal{U}_r$ is classified as inconsistent, \method re-optimizes $\mathcal{G}(1, b)$ using only $\mathcal{E}_o \cup \mathcal{C}$ to recover from a degraded estimate.
\section{Metrics}
\label{sec:metrics}
\subsection{On the Limitations of ATE and RPE for Robust SLAM Evaluation}
In the evaluation of SLAM systems, the estimated trajectory $\mathbf{x}^*$ is usually compared against the ground truth $\mathbf{x}^{gt}$ using two metrics: \textbf{Absolute Trajectory Error} (ATE) and \textbf{Relative Pose Error} (RPE)~\cite{zhang2018tutorial}.

ATE measures global trajectory accuracy by first estimating the rigid-body transformation $\mathbf{x}^{ATE} \in \acrshort{SE}(d)$ that best aligns the estimated and ground-truth trajectories. This transformation is then applied to the estimated trajectory for computing the position error between corresponding nodes. In contrast, the RPE measures local accuracy by comparing the relative transformations between consecutive poses in the two trajectories. Both metrics are reported as the root mean square error (RMSE).

In practice, ATE and RPE are widely used to compare robust optimizers for pose graph optimization, even though neither metric is specifically designed to evaluate robustness to outliers. As a result, their values can be heavily influenced by outlier-induced distortions, complicating the selection and tuning of robust solvers.
\begin{figure}[t]
    \centering
    \begin{minipage}[t]{.30\linewidth}
        \centering
        \includegraphics[width=\textwidth]{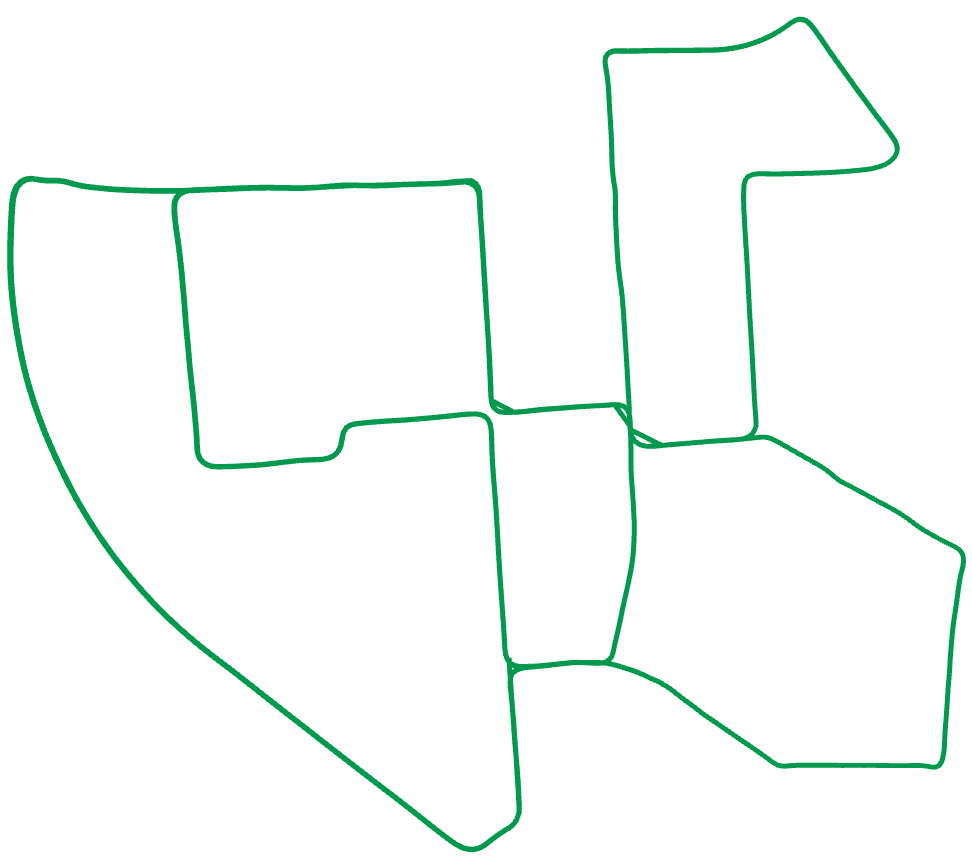}
        \center{\vspace*{-2ex}\footnotesize(a1) GT KITTI\_00}
    \end{minipage}
    \begin{minipage}[t]{.30\linewidth}
        \centering
        \includegraphics[width=\textwidth]{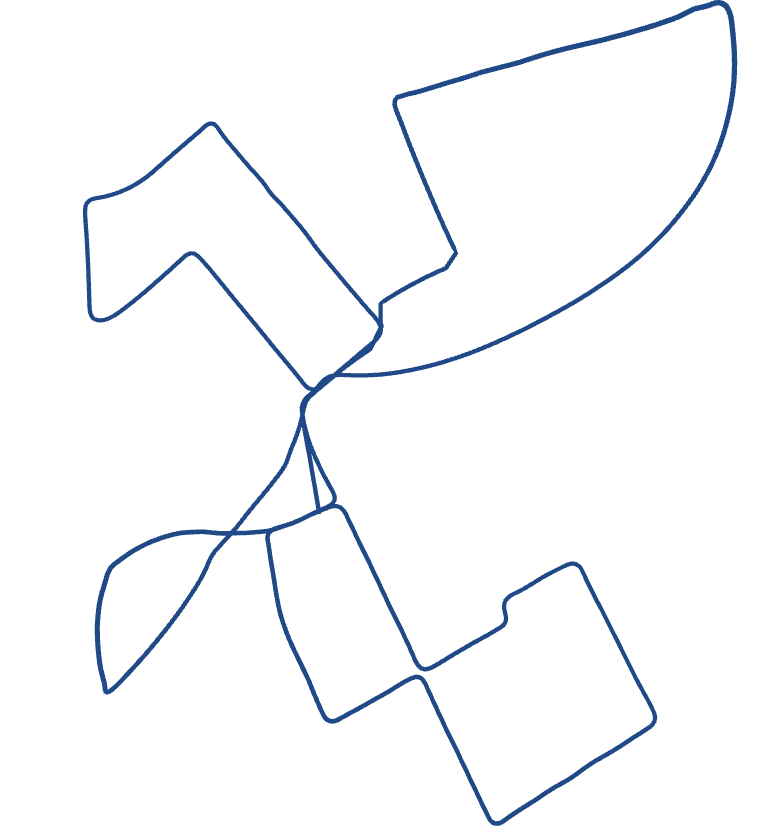}
        \center{\vspace*{-2ex}\footnotesize(a2) ATE=191.05 [m] RPE=3.48 [m] }
    \end{minipage}
    \begin{minipage}[t]{.30\linewidth}
        \centering
        \includegraphics[width=\textwidth]{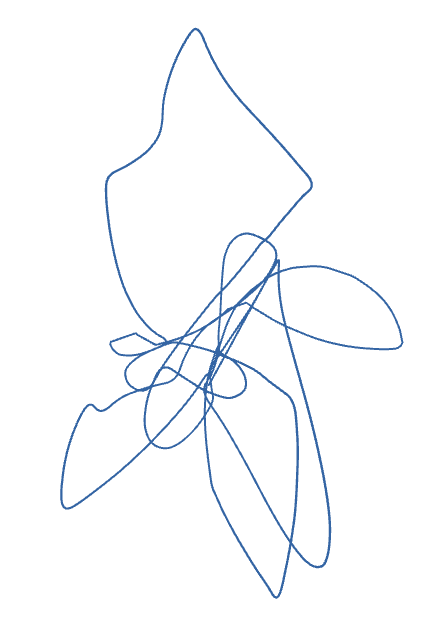}
        \center{\vspace*{-2ex}\footnotesize(a3) ATE=140.65 [m] RPE=0.39 [m]}
    \end{minipage}
    \begin{minipage}[t]{.30\linewidth}
        \centering
        \includegraphics[width=\textwidth]{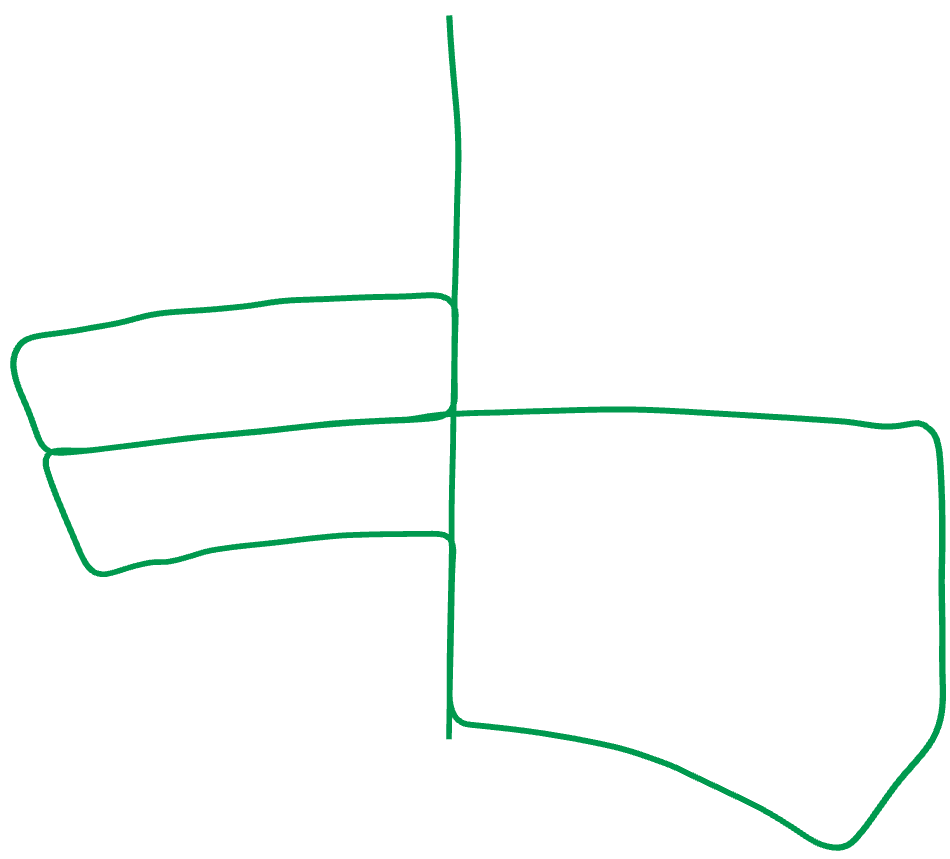}
        \center{\vspace*{-2ex}\footnotesize(b1) GT KITTI\_05}
    \end{minipage}
    \begin{minipage}[t]{.30\linewidth}
        \centering
        \includegraphics[width=\textwidth]{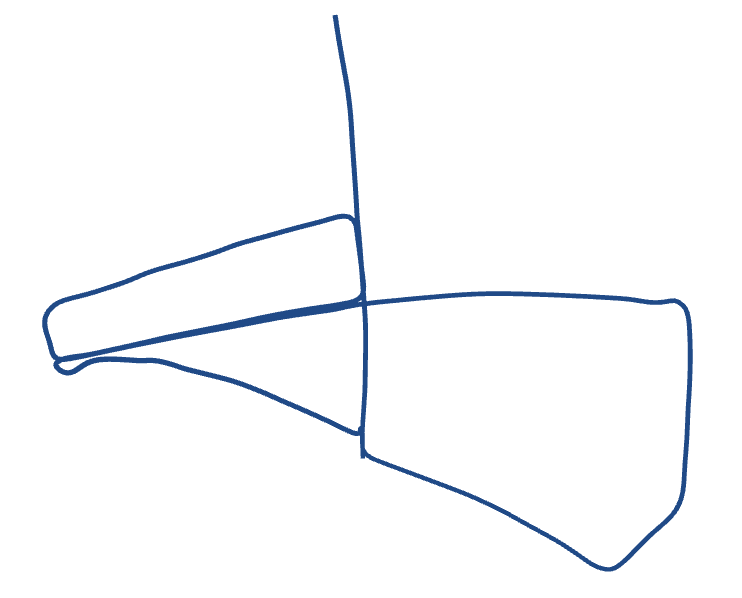}
        \center{\vspace*{-2ex}\footnotesize(b2) ATE=151.1 [m] RPE=2.66 [m]}
    \end{minipage}
    \begin{minipage}[t]{.30\linewidth}
        \centering
        \includegraphics[width=\textwidth]{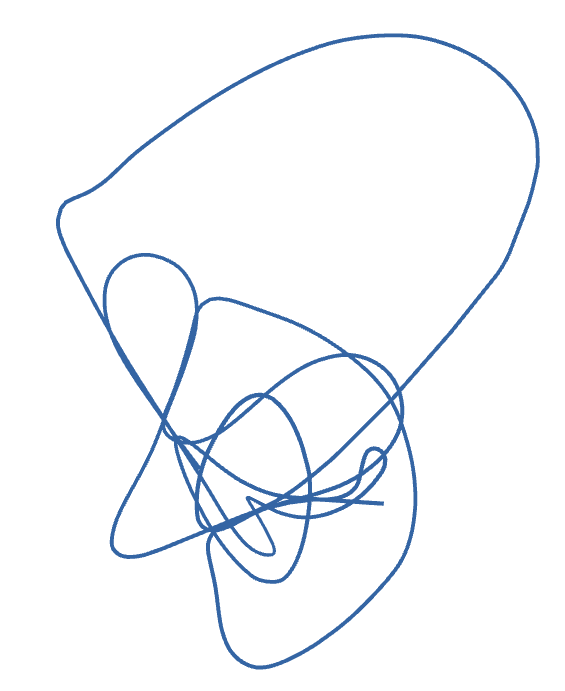}
        \center{\vspace*{-2ex}\footnotesize(b3) ATE=110.8 [m] RPE=0.25 [m]}
    \end{minipage}
    \caption{Examples of Absolute Trajectory Error (ATE) and Relative Pose Error (RPE) scores for trajectories under high distortion levels. Although trajectories (a2) and (b2) appear qualitatively superior, the ATE and RPE metrics indicate the opposite.}
    \label{fig:ate_rpe_no_sense}
\end{figure}

\figref{fig:ate_rpe_no_sense} illustrates examples of distorted trajectories alongside their corresponding ATE, RPE, and ground truth. In (a2) and (b2), the trajectories preserve their local structure, with (b2) remaining mostly unaltered, whereas (a3) and (b3) are severely distorted, making the ground truth trajectories in (a1) and (b1) difficult to recognize. From a qualitative standpoint, (a2) and (b2) are preferable to (a3) and (b3). However, ATE and RPE contradict this observation.

\begin{figure}[t]
    \centering
    \begin{minipage}[t]{.24\textwidth}
        \centering
        \includegraphics[width=\textwidth]{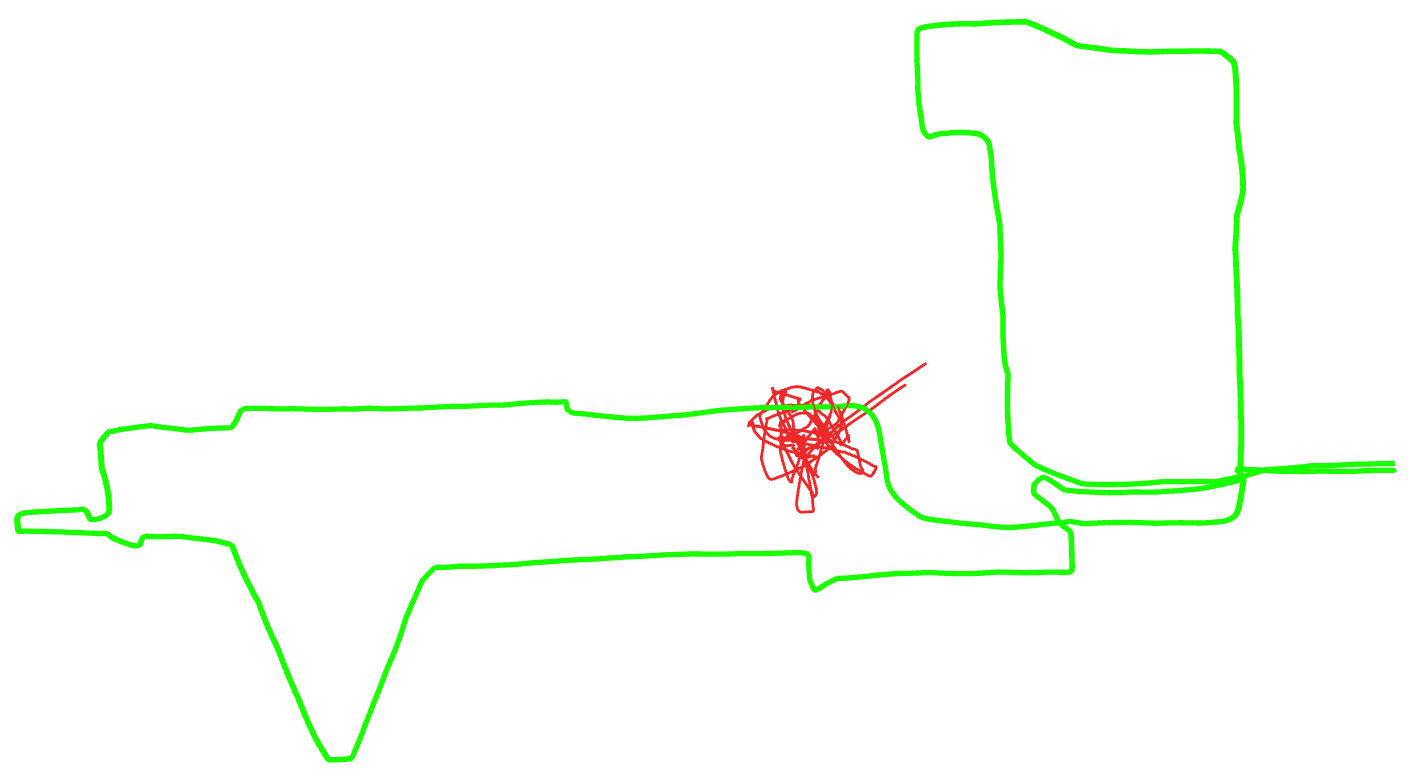}
        \center{\vspace*{-2ex}\footnotesize(a)}
    \end{minipage}
    \begin{minipage}[t]{.24\textwidth}
        \centering
        \includegraphics[width=\textwidth]{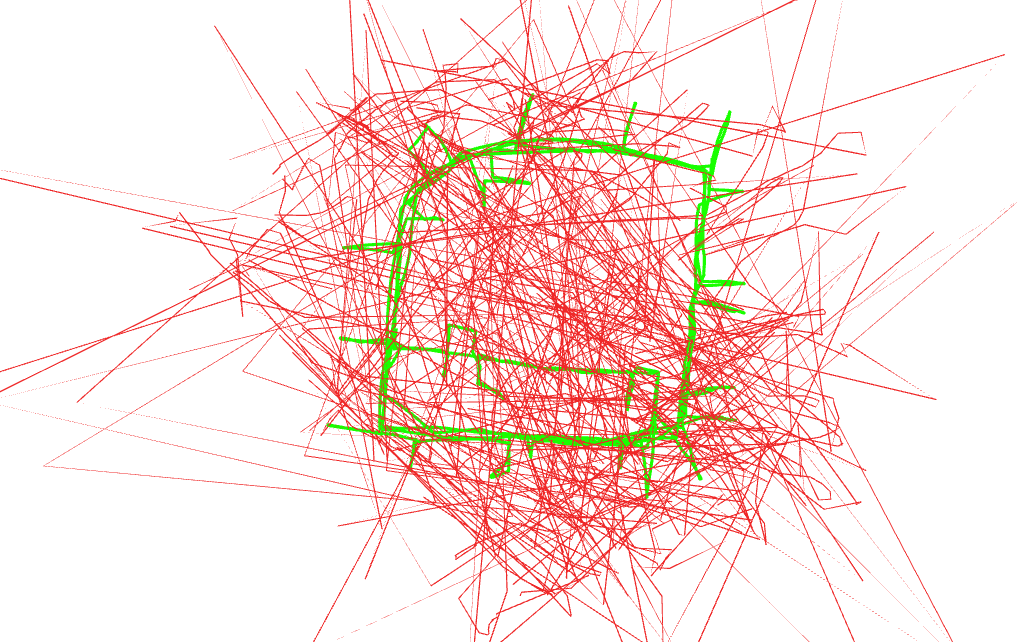}
        \center{\vspace*{-2ex}\footnotesize(b)}
    \end{minipage}
    \caption{Image representing two distortion effects that arise from incorrectly integrating outliers into the optimization process.}
    \label{fig:distortions}
\end{figure}

Failures in PGO can result in different types of trajectory distortions, as illustrated in \figref{fig:distortions}, with the following effects on ATE and RPE. In case (a), the trajectory collapses on itself, nearly degenerating into a point. The optimal ATE alignment, $\mathbf{x}^{ATE}$, moves the estimated trajectory $\mathbf{x}^*$ to the center of $\mathbf{x}^{gt}$, bounding the ATE to the average distance of the ground-truth poses $\mathbf{x}_i^{gt}$ from the trajectory’s center. Meanwhile, the RPE is bounded to the average displacement between consecutive ground-truth poses $\mathbf{x}_i^{gt}$ and $\mathbf{x}_{i+1}^{gt}$.
In case (b), severe disagreement between edges causes the trajectory to diverge outward due to numerical instability. Here, $\mathbf{x}^{ATE}$ aligns the ground truth $\mathbf{x}^{gt}$ with the center of the estimated trajectory $\mathbf{x}^*$, opposite to case (a). As a result, both ATE and RPE become unbounded.
   
\subsection{Metrics for RPGO}
ATE and RPE are meaningful metrics only when the estimated trajectory is not significantly distorted. Rather than relying directly on ATE, we use it to define the \textbf{Success Rate} (SR):  
\begin{equation*}
    SR = \frac{N_{suc}}{N_{tot}}
\end{equation*}
where $N_{suc}$ is the number of successful runs, and $N_{tot}$ is the total number of the runs. 

A run is deemed successful if the estimated trajectory achieves an ATE below a defined threshold $th_{ATE}$, otherwise, it is considered a failure due to trajectory distortion or insufficient drift correction. As a result, the SR reflects the likelihood of a robust optimizer converging to a good solution. This probabilistic interpretation provides a more reliable assessment than raw ATE values, which may fluctuate significantly and fail to reflect true performance, as illustrated in \figref{fig:ate_rpe_no_sense}.  

To provide further insight into the performance of the different \acrshort{RPGO} systems, we also report \textbf{Precision}, \textbf{Recall}, and \textbf{F1 score}:
\begin{equation*}
    pr = \frac{tp}{tp + fp}, \quad rc = \frac{tp}{tp + fn}, \quad F1 = 2 \cdot \frac{pr \cdot rc}{pr + rc}
\end{equation*}
The precision $pr$ captures how many outliers have been incorrectly integrated, while the recall $rc$ reflects how many inliers have been discarded. The F1 score combines both measures to provide an overall evaluation. Ideally, both precision and recall should be 1. However, low precision poses a greater risk than low recall, as it can result in a highly distorted trajectory. In contrast, low recall primarily sacrifices drift removal, which is less detrimental overall. 

Finally, we report the mean convergence time, defined as the average time required for the optimizer to converge after incorporating a new loop closure constraint.

\section{Experimental Setup}
\label{sec:experiments}

\subsection{Baseline methods}
We benchmark \acrshort{TACO} against six state-of-the-art \acrshort{RPGO} methods: MAXMIX~\cite{olson2013inference}, Switchable Constraints \cite{sunderhauf2012switchable}, Realizing-Reversing-Recovering \cite{latif2013robust}, Dynamic Covariance Scaling\cite{agarwal2013robust}, Huber~\cite{black1996unification}, and Graduated Non-Convexity~\cite{yang2020graduated}. To ensure fair comparison, all methods are implemented within the g2o framework~\cite{grisetti2011g2o}, avoiding discrepancies arising from differences between gtsam and g2o solvers. For Max-Mixture (MAXMIX)~\cite{olson2013inference} and Switchable Constraints (SC)~\cite{sunderhauf2012switchable}, we used the implementation provided in OpenSLAM\footnote{\url{https://openslam-org.github.io/}}. The Realizing-Reversing-Recovering (RRR) method~\cite{latif2013robust} is open-source and directly implemented in g2o. Dynamic Covariance Scaling (DCS)~\cite{agarwal2013robust} and Huber~\cite{black1996unification} are natively available in g2o, while Graduated Non-Convexity (GNC)~\cite{yang2020graduated} is implemented in g2o following the gtsam version~\cite{gtsam}.

\begin{figure*}[!t]
    \centering
    \includegraphics[width=0.99\linewidth]{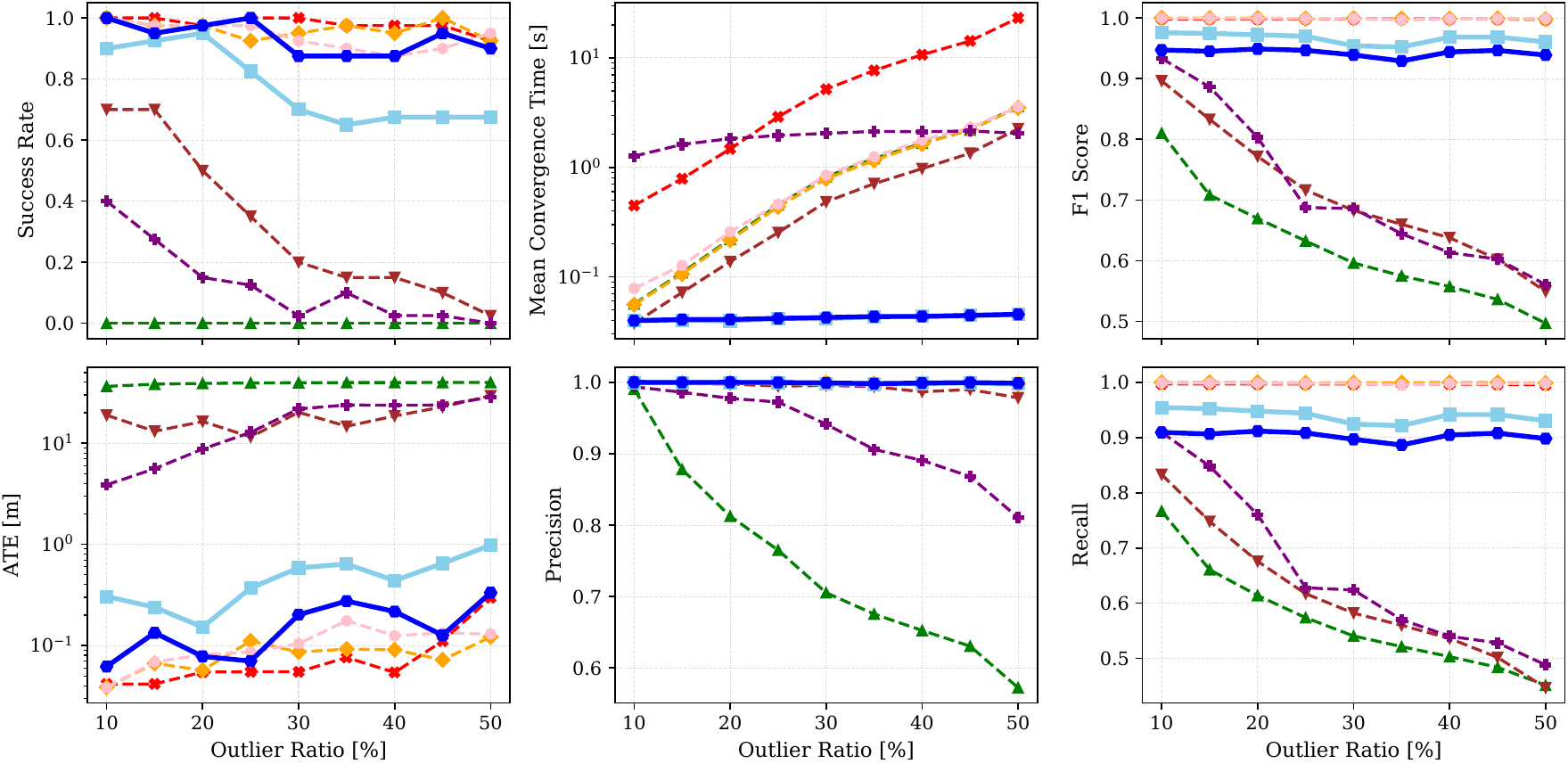}
    \caption{A comparison of the performance of \acrshort{TACO} and state-of-the-art methods on 2D SLAM datasets as the outlier percentage varies. The metrics are presented in the following order from left to right: Success Rate, Runtime, and F1 score. Legend of the graphs: \acrshort{TACO}(\protect\bluehexa), \acrshort{IPC}(\protect\lightbluesquare), HUBER~\cite{black1996unification}(\protect\greentriangle), DCS~\cite{agarwal2013robust}(\protect\goldendiamond), SC~\cite{sunderhauf2012switchable}(\protect\pinkcircle), MAXMIX~\cite{olson2013inference}(\protect\browntriangle), RRR~\cite{latif2013robust}(\protect\purplecross), GNC~\cite{yang2020graduated}(\protect\redx).}
    \label{fig:full_stats_2D}
\end{figure*}

\begin{table*}[!t] \centering
\begin{small}
\caption{Success rate (SR), precision (PR), and mean convergence time in ms (DT) for the compared methods on the 2D datasets as the outlier percentage varies. Bold indicates the best value per column.}
\label{tab:comparison_2D}
\footnotesize
\begin{tabular}{@{}l*{5}{ccc}@{}}\toprule
\multicolumn{1}{c}{\textbf{Method}} & \multicolumn{3}{c}{\textbf{10\%}} & \multicolumn{3}{c}{\textbf{20\%}} & \multicolumn{3}{c}{\textbf{30\%}} & \multicolumn{3}{c}{\textbf{40\%}} & \multicolumn{3}{c}{\textbf{50\%}} \\
\cmidrule(lr){2-4} \cmidrule(lr){5-7} \cmidrule(lr){8-10} \cmidrule(lr){11-13} \cmidrule(lr){14-16}
& SR $\uparrow$ & PR $\uparrow$ & DT $\downarrow$ & SR $\uparrow$ & PR $\uparrow$ & DT $\downarrow$ & SR $\uparrow$ & PR $\uparrow$ & DT $\downarrow$ & SR $\uparrow$ & PR $\uparrow$ & DT $\downarrow$ & SR $\uparrow$ & PR $\uparrow$ & DT $\downarrow$ \\ \midrule
\textbf{GNC} & \textbf{100.0} & \textbf{100.0} & 449.0 & \textbf{97.5} & 99.9 & 1474.3 & \textbf{100.0} & \textbf{99.9} & 5171.9 & \textbf{97.5} & \textbf{99.9} & 10690.5 & 92.5 & \textbf{99.9} & 23159.1 \\ \hdashline
\textbf{DCS} & \textbf{100.0} & 99.9 & 55.7 & \textbf{97.5} & 99.9 & 216.2 & 95.0 & 99.8 & 785.8 & 95.0 & 99.8 & 1635.9 & 92.5 & 99.7 & 3502.2 \\ \hdashline
\textbf{MAXMIX} & 70.0 & 99.9 & \textbf{37.6} & 50.0 & 99.7 & 137.7 & 20.0 & 99.5 & 484.5 & 15.0 & 98.6 & 974.1 & 2.5 & 97.8 & 2253.3 \\ \hdashline
\textbf{SC} & \textbf{100.0} & \textbf{100.0} & 78.3 & \textbf{97.5} & 99.9 & 257.8 & 92.5 & \textbf{99.9} & 847.0 & 87.5 & 99.8 & 1778.5 & \textbf{95.0} & \textbf{99.9} & 3577.8 \\ \hdashline
\textbf{RRR} & 40.0 & 99.3 & 1265.3 & 15.0 & 97.7 & 1830.2 & 2.5 & 94.1 & 2045.0 & 2.5 & 89.0 & 2115.7 & 0.0 & 81.0 & 2043.5 \\ \hdashline
\textbf{IPC\_S10} & 90.0 & 99.9 & 39.8 & 95.0 & 99.9 & \textbf{39.6} & 70.0 & 99.8 & \textbf{41.4} & 67.5 & 99.8 & \textbf{43.5} & 67.5 & 99.7 & \textbf{45.4} \\ \hdashline
\textbf{TACO} & \textbf{100.0} & \textbf{100.0} & 39.8 & \textbf{97.5} & \textbf{100.0} & 40.8 & 87.5 & \textbf{99.9} & 42.5 & 87.5 & \textbf{99.9} & \textbf{43.5} & 90.0 & 99.8 & 45.5 \\
\bottomrule
\end{tabular}
\end{small}
\end{table*}

\subsection{Datasets}
The datasets used for evaluation are divided into two categories: 2D and 3D. The 2D datasets, described in~\cite{carlone2014fast, carlone2014angular}, are publicly available and include Intel, Csail, Freiburg Building (FR079), and Freiburg University Hospital (FRH), all of which originate from real-world measurements.

For the Visual SLAM evaluation, real-world datasets had to be adapted to the \acrshort{RPGO} task. KITTI\_00 and KITTI\_05 were derived from sequences 00 and 05 of KITTI's visual odometry dataset~\cite{Geiger2012CVPR}, while TUM\_FR1\_DESK was extracted from the corresponding sequence in TUM's RGB-D SLAM Dataset~\cite{sturm12iros}. Odometry and loop closure constraints were obtained using ORB\_SLAM3~\cite{9440682} on the raw data from these sequences. 

The loop closure edges in the unaltered datasets served as ground truth for inliers.  A pseudo-ground-truth trajectory was generated using SE-Sync~\cite{rosen2019se}, a fast non-minimal solver for PGO, applied to the outlier-free version of each dataset. 

Outliers were artificially added using the Vertigo package~\cite{vertigo} as a percentage of the total loop closures to ensure balanced evaluation. Outliers are generated using the random constraint policy, which introduces constraints between randomly selected poses.

Each dataset includes nine difficulty levels, representing outlier ratios from 10\% to 50\% in 5\% increments. 
The 50\% level represents the extreme case in which outliers are equal in number to inliers, meaning that the loop closure detector is as likely to generate an incorrect correspondence as a correct one, which is unrealistic in real-world applications. For each level, ten independent corrupted instances are generated.

\subsection{Implementation details}
\method~is implemented in C++, using the g2o framework~\cite{grisetti2011g2o} and the original Switchable Constraints implementations~\cite{sunderhauf2012switchable}. Optimization is performed via Gauss-Newton~\cite{nocedal1999numerical}. \acrshort{IPC} is configured with $s = 10$ and \acrshort{SOS} with $M = 10$ unless otherwise mentioned, with ablation studies provided in~\secref{effect_s} and~\secref{effect_m}. 

Parameters are kept constant across datasets with the same dimensionality (separate settings are used for 2D and Visual SLAM). Experiments are conducted on a PC with a single Intel(R) Xeon(R) Gold 5220 CPU. Following~\cite{mcgann2023robust}, methods that do not explicitly classify measurements treat a loop closure as an inlier if its $\chi^2$ error is below $\chi^2_{th}$ with $\alpha_{th} = 0.95$.

To simulate incremental operation, loop closure detections are provided sequentially in chronological order, one per step. Performance is evaluated at the end of the sequence. For success rate (SR) evaluation, we use $th_{ATE} = 0.75$\,m, and compute ATE and RPE using the libraries from~\cite{guadagnino2021hipe}.

Except for the methods that explicitly provide parameter settings, all remaining methods, including \acrshort{IPC} and \acrshort{TACO}, were configured through a grid search. The search was carried out on one representative dataset per dimensionality (Intel for 2D, KITTI\_00 for 3D), each corrupted with 0\% (original), 5\%, and 10\% outliers. The parameter configuration achieving the best average SR across these three levels was selected.

\section{Results}
\subsection{Comparison with State-of-the-art Methods on 2D Datasets}
The following results evaluate robustness of the methods across increasing outlier ratios, supporting claims 1 and 2: that \acrshort{IPC} provides an effective online consistent-set approximation, and that \acrshort{SOS} recovers from residual errors while preserving runtime efficiency.

\figref{fig:full_stats_2D} reports performance metrics, including success rate, mean convergence time, F1 score, ATE, precision, and recall, averaged over all 2D datasets. \tabref{tab:comparison_2D} complements these results by summarizing the main metrics, providing a clearer comparison where visual trends overlap.

In terms of SR, \acrshort{TACO} achieves around $90\%$ across all outlier percentages, with performance comparable to \emph{offline} state-of-the-art methods such as GNC, SC, and DCS. Among these, GNC achieves the highest SR, exceeding $97\%$ in most cases. \acrshort{IPC} maintains SR above $80\%$ up to $25\%$ outliers, after which it drops to around $65\%$. Adding \acrshort{SOS} to \acrshort{IPC} increases SR by approximately $15\%$ on average and ensures consistent performance across all outlier levels.

While achieving SR comparable to GNC, with only a $5\%$ average difference, \acrshort{TACO} has a mean convergence time of just 45\,ms, corresponding to an average speedup of about $150\times$ compared to GNC. As shown in \tabref{tab:comparison_2D}, methods such as GNC, SC, DCS, MAXMIX, and HUBER exhibit increasing convergence time with higher outlier ratios. For example, MAXMIX's convergence time (which is the fastest comparison method) increases from 37.6\,ms for 10\% outliers to over 2.2\,s for 50\% outliers. This compares to a near constant runtime of \acrshort{TACO} (from 39.8\,ms to 45.5\,ms) and RRR (from 1.2\,s to 2.0\,s). We note that the introduction of \acrshort{SOS} has minimal impact on runtime, keeping \acrshort{TACO} close to \acrshort{IPC}, with only a negligible overhead of about $0.009\times$ on average.

Precision is the metric that most strongly affects SR, and should ideally be close to one, since even small amounts of incorrectly accepted outliers can severely degrade the trajectory estimate. Both the quantity and the quality of accepted outliers are important for successful convergence. As shown in \tabref{tab:comparison_2D}, even a $0.1\%$ difference in precision between SC at $30\%$ and $40\%$ outlier levels results in a $5\%$ drop in SR. The top-performing methods, namely GNC, DCS, SC, and \acrshort{TACO}, consistently maintain precision at or above $99.7\%$. In contrast, the poor SR of RRR and HUBER is attributable to low precision. RRR shows a clear degradation in both precision and SR as the outlier ratio increases, from an SR of $40\%$ and a precision of $99.3\%$ at $10\%$ outliers to an SR of $0\%$ and a precision of $81\%$ at $50\%$, while HUBER consistently fails to converge, yielding an SR of $0\%$ across all scenarios.

Recall mainly affects drift removal and therefore also influences SR, but less strongly than precision. GNC, DCS, and SC achieve recall close to $100\%$, while \acrshort{IPC} reaches around $95\%$ and \acrshort{TACO} around $90\%$. Despite its lower recall, \acrshort{TACO} achieves higher SR than \acrshort{IPC} due to its improved precision. Conversely, MAXMIX’s poor SR is due to excessive inlier rejection, which limits its ability to mitigate drift.

The F1 score provides a global view of performance by combining precision and recall, but it can be misleading since it assigns equal weight to both metrics. For example, MAXMIX and RRR show similar F1 trends, even though MAXMIX achieves higher SR than RRR. Likewise, \acrshort{IPC} has a higher F1 score than \acrshort{TACO}, despite \acrshort{TACO} achieving better SR. Thus, for the evaluation of RPGO systems, it is more informative to consider precision and recall separately, as this provides clearer insight into whether failures arise from incorrect outlier acceptance or insufficient drift removal.

To further support the claim that ATE alone is not a suitable metric for evaluating \acrshort{RPGO} systems, \figref{fig:full_stats_2D} shows that ATE would incorrectly suggest RRR performs better than MAXMIX, whereas SR clearly indicates the opposite. This discrepancy arises because when MAXMIX fails to converge, its estimated trajectory diverges significantly, while RRR also diverges but less severely.

\begin{table*}[!t] 
\centering
\caption{Success rate (SR), precision (PR), and mean convergence time in ms (DT) for the compared methods on the 3D datasets as the outlier percentage varies. Bold indicates the best value per column.}
\label{tab:comparison_3D}
\begin{small}
\footnotesize
\setlength{\tabcolsep}{5pt}
\begin{tabular}{@{}l*{5}{ccc}@{}}\toprule
\multicolumn{1}{c}{\textbf{Method}} & \multicolumn{3}{c}{\textbf{10\%}} & \multicolumn{3}{c}{\textbf{20\%}} & \multicolumn{3}{c}{\textbf{30\%}} & \multicolumn{3}{c}{\textbf{40\%}} & \multicolumn{3}{c}{\textbf{50\%}} \\
\cmidrule(lr){2-4} \cmidrule(lr){5-7} \cmidrule(lr){8-10} \cmidrule(lr){11-13} \cmidrule(lr){14-16}
& SR $\uparrow$ & PR $\uparrow$ & DT $\downarrow$ & SR $\uparrow$ & PR $\uparrow$ & DT $\downarrow$ & SR $\uparrow$ & PR $\uparrow$ & DT $\downarrow$ & SR $\uparrow$ & PR $\uparrow$ & DT $\downarrow$ & SR $\uparrow$ & PR $\uparrow$ & DT $\downarrow$ \\ \midrule
\textbf{GNC} & \textbf{100.0} & \textbf{100.0} & 467.3 & 80.0 & 86.6 & 754.0 & 76.6 & 83.3 & 1133.9 & 80.0 & 83.3 & 1016.3 & 43.3 & 53.3 & 936.0 \\ \hdashline
\textbf{DCS} & \textbf{100.0} & \textbf{100.0} & 123.1 & 90.0 & \textbf{100.0} & 116.4 & 96.6 & \textbf{100.0} & 123.0 & \textbf{100.0} & \textbf{100.0} & 128.1 & \textbf{100.0} & \textbf{100.0} & 137.4 \\ \hdashline
\textbf{RRR} & 90.0 & 96.6 & 2983.4 & 83.3 & 89.8 & 5687.6 & 96.6 & 95.0 & 8204.1 & 66.6 & 91.3 & 10920.3 & 63.3 & 61.6 & 14142.3 \\ \hdashline
\textbf{MAXMIX} & 53.3 & 86.6 & \textbf{74.6} & 46.6 & 83.0 & \textbf{81.7} & 33.3 & 69.0 & \textbf{87.6} & 36.6 & 68.8 & \textbf{100.0} & 26.6 & 58.5 & \textbf{110.1} \\ \hdashline
\textbf{SC} & \textbf{100.0} & \textbf{100.0} & 133.4 & 93.3 & \textbf{100.0} & 135.8 & \textbf{100.0} & \textbf{100.0} & 145.6 & \textbf{100.0} & \textbf{100.0} & 153.0 & 96.6 & \textbf{100.0} & 162.7 \\ \hdashline
\textbf{TACO} & 96.6 & \textbf{100.0} & 93.8 & \textbf{96.6} & \textbf{100.0} & 95.0 & 83.3 & 99.9 & 112.4 & 90.0 & 99.8 & 105.0 & 83.3 & \textbf{100.0} & 120.2 \\
\bottomrule
\end{tabular}
\end{small}
\end{table*}

\subsection{Comparison with State-of-the-art Methods on 3D Datasets}
\begin{figure}[!t]
    \centering
    \includegraphics[width=0.99\linewidth]{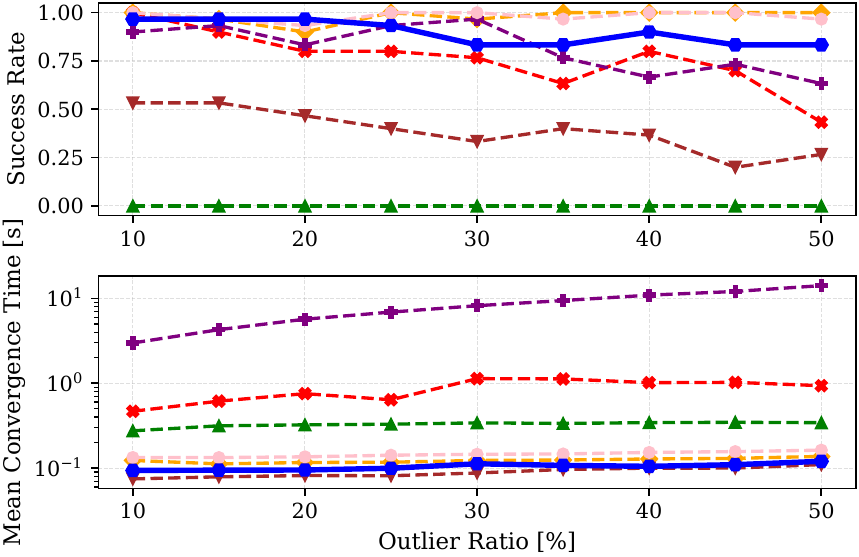}
    \caption{A comparison of the performance of \acrshort{TACO} and state-of-the-art methods on 3D Visual SLAM datasets as the outlier percentage varies. The metrics are presented in the following order from top to bottom: Success Rate and Mean Convergence Time. Legend of the graphs: \acrshort{TACO}(\protect\bluehexa), HUBER~\cite{black1996unification}(\protect\greentriangle), DCS~\cite{agarwal2013robust}(\protect\goldendiamond), SC~\cite{sunderhauf2012switchable}(\protect\pinkcircle), MAXMIX~\cite{olson2013inference}(\protect\browntriangle), RRR~\cite{latif2013robust}(\protect\purplecross), GNC~\cite{yang2020graduated}(\protect\redx).}
    \label{fig:succ_dt_3D}
\end{figure}

This experiment repeats the main comparison on the 3D Visual SLAM datasets, namely KITTI\_00, KITTI\_05, and TUM\_FR1\_DESK, to evaluate whether the performance trends from the 2D case generalize to 3D. \figref{fig:succ_dt_3D} reports only success rate and mean convergence time, as the other metrics are discussed in the 2D case. 

\begin{figure*}[t!]
    \centering
    \includegraphics[width=0.99\linewidth]{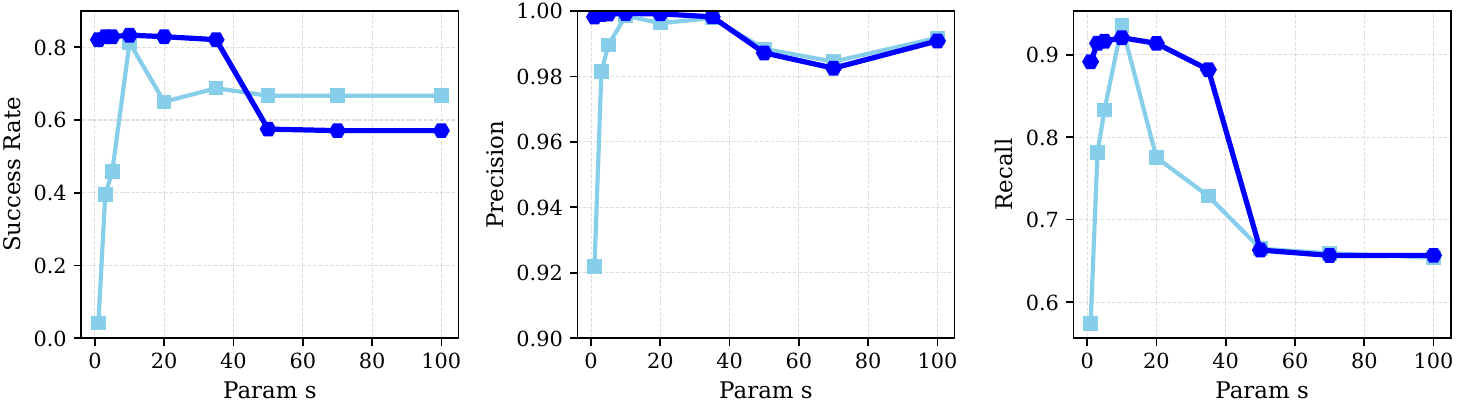}
    \caption{Comparison of \acrshort{IPC} and \acrshort{TACO} performance on 2D SLAM datasets with varying parameter $s$. Results are averaged across all datasets and outlier percentages. The metrics are presented from left to right: Success Rate, Precision, and Recall. Legend of the graphs: \acrshort{TACO}(\protect\bluehexa), \acrshort{IPC}(\protect\lightbluesquare).}
    \label{fig:effect_S}
\end{figure*}

\begin{figure*}[t!]
    \centering
    \includegraphics[width=0.99\linewidth]{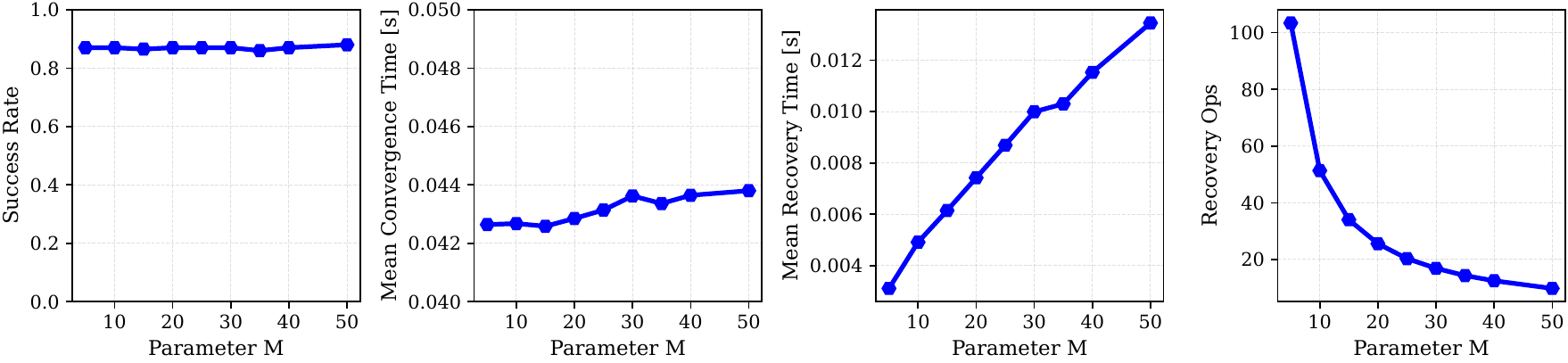}
    \caption{Evaluation of \acrshort{TACO} (\protect\bluehexa) performance on 2D SLAM datasets with varying parameter $M$. Results are averaged across all datasets and outlier percentages. The metrics, presented from left to right, are Success Rate, Mean Convergence Time, Mean Recovery Time, and Recovery Operations.}
    \label{fig:effect_M}
\end{figure*}

The increase in dimensionality significantly affects GNC, whose SR is now outperformed by DCS and SC, both achieving values exceeding $90\%$, as well as by \method, which consistently maintains SR above $83\%$. RRR and MAXMIX exhibit decreasing trends, with SR remaining above $63\%$ and $26\%$. In this setting, RRR performs substantially better than in the 2D case, while MAXMIX exhibits behavior consistent with the trends observed in the 2D experiments.

In terms of runtime, \method maintains an approximately constant mean convergence time of 100\,ms across all outlier levels. DCS and SC achieve mean convergence times of approximately 120\,ms and 140\,ms, respectively, while MAXMIX is the fastest method, averaging approximately 90\,ms. Since the computational complexity of the offline methods scales with the number of loop closures, and only a limited number of true loop closures are detected ($\approx$60) due to the conservative ORB-SLAM3 loop detector, their overall runtime is reduced. Nevertheless, RRR and GNC remain the most computationally expensive methods.

\subsection{Comparison with State-of-the-art Methods on Extreme Outlier Rates}
\label{extreme_out}
For completeness, we conduct an ablation study under extreme outlier rates exceeding 50\% to evaluate method performance in highly challenging conditions. For the Intel, FR079, and Csail datasets, four additional difficulty levels are generated with outlier ratios from 60\% to 90\% in 10\% increments.

\tabref{tab:extreme_outlier_rate} reports success rate and mean convergence time for the best-performing methods from the 2D evaluation, including \method.

The results show that as the deviation from assumption [\textbf{A2}], one of the core principles of \acrshort{TACO}, performance degrades accordingly, with SR decreasing from 80\% at 60\% to 53\% at 90\%. SC achieves the best overall performance, reaching 100\% SR at 60\% and 70\%, and remaining above 86\% SR thereafter. It is followed by DCS and GNC, which, like \acrshort{TACO}, exhibit a decreasing trend: DCS drops from 93\% SR at 60\% to 60\% SR at 90\%, and GNC from 80\% SR to 63\%.

While GNC and \acrshort{TACO} differ by approximately 7\% in SR on average, \acrshort{TACO} maintains a mean convergence time below 40\,ms, whereas GNC mean convergence time increases significantly from 1.2\,s to 67\,s. 
A similar trend is observed for DCS and SC, whose mean convergence time increases from 180\,ms to 9.2\,s.

Although \acrshort{TACO} is designed for more realistic outlier regimes, it still achieves a reasonable SR of 53\%, only about 10 percentage points below more complex methods such as GNC and DCS, while providing a substantial improvement in runtime.

\begin{table}[!t]
\centering
\begin{small}
\footnotesize
\setlength{\tabcolsep}{3pt}
\renewcommand{\arraystretch}{1.1}
\begin{tabular}{@{}l*{4}{cc}@{}}
\toprule
\multicolumn{1}{c}{\textbf{Method}} &
\multicolumn{2}{c}{\textbf{60\%}} &
\multicolumn{2}{c}{\textbf{70\%}} &
\multicolumn{2}{c}{\textbf{80\%}} &
\multicolumn{2}{c}{\textbf{90\%}} \\
\cmidrule(lr){2-3}
\cmidrule(lr){4-5}
\cmidrule(lr){6-7}
\cmidrule(lr){8-9}
& SR $\uparrow$ & DT $\downarrow$
& SR $\uparrow$ & DT $\downarrow$
& SR $\uparrow$ & DT $\downarrow$
& SR $\uparrow$ & DT $\downarrow$ \\
\midrule
\textbf{GNC}  & 80.0 & 1436.8 & 80.0 & 3111.0 & 70.0 & 10758.0 & 63.3 & 67410.6 \\ \hdashline
\textbf{DCS}  & 93.3 & 176.0 & 86.6 & 383.8 & \textbf{90.0} & 1438.6 & 60.0 & 9239.0 \\ \hdashline
\textbf{SC}   & \textbf{100.0} & 186.8 & \textbf{100.0} & 414.0 & \textbf{90.0} & 1482.4 & \textbf{86.6} & 9389.0 \\ \hdashline
\textbf{TACO} & 80.0 & \textbf{39.7} & 70.0 & \textbf{35.2} & 63.3 & \textbf{36.0} & 53.3 & \textbf{33.9} \\
\bottomrule
\end{tabular}
\end{small}
\caption{Success rate (SR), and mean convergence time in ms (DT) for the compared methods on the 2D datasets for extreme outlier rates. Bold indicates the best value per column.}
\label{tab:extreme_outlier_rate}
\end{table}

\subsection{Effect of the parameter \texorpdfstring{$s$}{s}}
\label{effect_s}
This section analyzes the impact of the parameter $s$ on \acrshort{IPC} and \acrshort{TACO}, with $M = 10$ fixed for \acrshort{TACO}. Experiments are conducted on the Intel, FR079, and Csail datasets, and results are averaged across all datasets and outlier levels up to 50\%.

As shown in \figref{fig:effect_S}, \acrshort{IPC} is particularly sensitive to $s$ in the range $[0, 20]$, reaching its highest SR at $s = 10$. In contrast, \acrshort{TACO} remains stable over a large range $[0, 50)$, maintaining an SR above $80\%$ thanks to its sanitization step, which compensates for the overly permissive behavior of \acrshort{IPC} at low $s$.

When $s > 50$, the performance of the two methods converges, with \acrshort{TACO} exhibiting slightly lower SR, precision, and recall. In this regime, the sanitization step begins to remove inliers from the consensus set $\mathcal{C}$, reflecting an overly conservative behavior of \acrshort{SOS} in poorly constrained scenarios, primarily due to the high rejection rate introduced by \acrshort{IPC}.

The inclusion of \acrshort{SOS} enables \method to maintain stable performance across a wide range of $s$ values, significantly reducing the complexity of parameter tuning \acrshort{IPC}.

\subsection{Effect of the parameter \texorpdfstring{$M$}{M}}
\label{effect_m}
This section analyzes the impact of the parameter $M$ on \acrshort{TACO} performance. Experiments are conducted on all 2D datasets, and results are averaged across all datasets and outlier levels up to 50\%. In \figref{fig:effect_M}, we denote the execution time of \acrshort{SOS} as the mean recovery time, and the number of executions per run as the number of recovery operations.  

As shown in \figref{fig:effect_M}, the SR consistently remains above $80\%$, as \acrshort{SOS} reliably recovers from degenerate solutions whenever it is invoked. The parameter $M$ primarily controls the trade-off between mean recovery time and the number of recovery operations.

For small values of $M$, recovery operations occur frequently, but each operates on a small trusted subgraph $\mathcal{G}^T$, resulting in low mean recovery time. For larger $M$, recovery is triggered less often, but each execution involves a larger subgraph and thus higher mean recovery time. These opposing effects balance out, leading to a nearly constant mean convergence time of approximately 43\,ms. Therefore, $M$ does not significantly affect the final estimate, but it governs the system’s reactivity to degraded solutions.
\section{Conclusions}
\label{sec:conclusions}
We presented \method, a two-stage framework that combines incremental consistent-set maximization (\acrshort{IPC}) with retrospective outlier sanitization (\acrshort{SOS}) to achieve robust pose graph optimization in an online setting. Evaluated on 2D and 3D Visual SLAM benchmarks, \method achieves success rates above $90\%$ and $83\%$, respectively, across outlier rates up to 50\% and without requiring any parameter retuning, while maintaining mean convergence times of approximately 45\,ms in 2D and 100\,ms in 3D. These results show that the test-and-check decomposition effectively bridges the gap between online performance and the robustness of offline state-of-the-art approaches.

\method's test component, \acrshort{IPC}, evaluates the measurement consistency of each loop closure measurement on the corresponding independent subgraph. By reweighting the cost function to emphasize odometry constraints and performing localized consistency checks, \acrshort{IPC} enables consistency evaluation while maintaining performance.

\method's check component, \acrshort{SOS}, leverages the switchable constraints to identify and remove outliers that may have been deemed consistent by \acrshort{IPC}. 
Its execution frequency is controlled by the parameter $M$, which determines how frequently past decisions are revisited, providing a mechanism to balance responsiveness and computational cost.

Several directions for future work can be identified. First, strategies inspired by Graduated Non-Convexity could be explored to automatically tune the scaling parameter $s$, or more generally to design adaptive mechanisms for its dynamic selection. Second, the development of heuristics aimed at maximizing the connectivity of trusted subgraphs may further improve the convergence behavior of \acrshort{SOS}.

Another promising direction for future work involves developing more advanced recovery mechanisms that reduce reliance on the initial estimate produced by \acrshort{IPC}, enabling more effective performance at extreme outlier rates. In particular, extending \acrshort{SOS} to recover inliers that were erroneously rejected, in addition to sanitizing incorrectly accepted outliers, is promising.

Moreover, it would be valuable to analyze the conditions under which \acrshort{IPC} guarantees reliable performance, for example in relation to graph size and measurement uncertainty. Exploring alternative graph parameterizations, such as cycle basis representations, may also provide deeper insights into theoretical performance guarantees. Finally, investigating the use of Gaussian Belief Propagation could enable the design of active mechanisms for outlier detection within the optimization process.

\section*{Author contributions}
\textbf{Dr.~Emilio Olivastri}: Investigation, Methodology, Formal Analysis, Data curation, Writing – original draft.\\
\textbf{Prof.~Alberto Pretto}: Conceptualization, Supervision, Writing – review \& editing.\\ 
\textbf{Prof.~Tobias Fischer}: Conceptualization, Supervision, Funding acquisition, Project administration, Writing – review \& editing.

\section*{Statements and declarations}

\subsection*{Declaration of conﬂicting interests}
The authors declared no potential conflicts of interest with respect to the research, authorship, and/or publication of this article.

\subsection*{Funding}
This research was partially supported by funding from ARC DECRA Fellowship DE240100149 to TF and the QUT Centre for Robotics.

\subsection*{Ethical considerations}
Not applicable.

\subsection*{Consent to participate}
Not applicable.

\subsection*{Consent for publication}
Not applicable.

\bibliographystyle{SageV}
\bibliography{references.bib}

\end{document}